\documentclass{article}
\usepackage{graphicx} 
\usepackage{authblk}
\usepackage{multirow}
\usepackage{amssymb}
\usepackage[table]{xcolor}\usepackage{tikz}
\usepackage{tcolorbox}

\usepackage{pifont} 
\usepackage{amsmath} 
\usepackage{makecell}
\usepackage{fvextra} 
\usepackage{multirow}
\usepackage{booktabs}
\usepackage[colorlinks=true,
            linkcolor=blue,
            citecolor=blue,
            urlcolor=blue]{hyperref}        
\usepackage{longtable}  
\usepackage[font=small]{caption}

\usepackage{array} 
\usepackage{float} 
\usepackage{caption}
\usepackage{xcolor}
\usepackage{comment}
\usepackage{hyperref}
\usepackage[numbers,sort&compress]{natbib}
\usepackage{url}
\usepackage[a4paper, total={7in, 9in}]{geometry}
\usepackage{lipsum} 
\newcommand{\keywords}[1]{\textbf{Keywords:} #1}
\newcommand{\hide}[1]{}

\title{AppleScab-LT: A Longitudinal Real-Field Apple Scab Dataset for Temporal Disease Progression Analysis}

\author{
\small
\textit{Aamir Hilal}$^{1*}$ \quad
\textit{Shabir Ahmad Sofi}$^{1}$ \quad
\textit{Neeraj Goel}$^{2}$ \\
\vspace{-10pt}
{\footnotesize
\textcolor{blue}{aamir\_hilal@nitsri.ac.in} \quad
\textcolor{blue}{shabir@nitsri.ac.in} \quad
\textcolor{blue}{neeraj@iitrpr.ac.in}
}
}

\date{
\vspace{-3mm}
{\footnotesize
$^{1}$Department of Information Technology, National Institute of Technology, Hazratbal Srinagar 190006, Jammu \& Kashmir, India \\
$^{2}$Department of Computer Science and Engineering, Indian Institute of Technology, Ropar Rupnagar 140001, Punjab, India
$^{*}$\textbf{Corresponding author:} \textcolor{blue}{aamir\_hilal@nitsri.ac.in}
}
}
\begin{document}
\maketitle
\vspace{-8mm}
\begin{abstract}
The development of reliable artificial intelligence systems for plant disease monitoring is constrained by the limited availability of longitudinal datasets that capture disease progression under natural field conditions. Although numerous plant disease datasets have advanced image-based disease recognition, most comprise static images acquired at a single time point, limiting their suitability for studying temporal disease evolution, severity progression, and temporal disease progression analysis. To address this gap, this study presents \textbf{AppleScab-LT}, a longitudinal real-field dataset specifically developed for monitoring apple scab disease progression through repeated observations of individually tracked infected leaves. Guided by a research-question-driven framework, the dataset was systematically developed, validated, and characterized to support reliable longitudinal disease analysis. AppleScab-LT was constructed through systematic orchard monitoring under natural environmental conditions. The dataset development pipeline includes longitudinal leaf tracking, expert-guided disease verification, polygon-based annotation, leaf isolation, disease severity quantification, and temporal sequence construction to accurately capture disease evolution. To ensure scientific reliability, a comprehensive quality assurance framework incorporating standardized annotation protocols, expert validation, automated integrity checks, sequence-level verification, and temporal consistency analysis was employed throughout the dataset curation process. The resulting dataset comprises \textbf{21 longitudinal leaf sequences}, containing \textbf{2,101 high-resolution images} and \textbf{264 progressive temporal samples} generated from repeated monitoring of the same infected leaves. AppleScab-LT captures substantial variability in disease progression, including differences in severity accumulation, progression rates, monitoring duration, and inter-leaf variability. Furthermore, quantitative disease descriptors based on pixel severity, color-intensity severity, and normalized relative severity provide standardized measurements for temporal disease analysis. By addressing the scarcity of longitudinal plant disease datasets, AppleScab-LT provides a reliable resource for advancing temporal disease intelligence, disease progression modelling, precision agriculture, and next-generation crop health monitoring systems.
\end{abstract}


\section{Introduction}
\label{sec:Introduction}

Plant diseases continue to pose a significant threat to global agricultural productivity, resulting in substantial yield losses, deterioration in crop quality, and considerable economic damage worldwide. Among horticultural crops, apple production is particularly vulnerable to fungal diseases, with apple scab (\textit{Venturia inaequalis}) being one of the most economically destructive diseases affecting commercial orchards worldwide \cite{bowen2011venturia}. The disease infects leaves, fruits, blossoms, and young shoots, causing premature defoliation, reduced fruit quality, decreased market value, and increased production costs due to repeated fungicide applications. Since disease severity can increase rapidly under favourable environmental conditions, continuous monitoring of disease development is essential for timely intervention and effective orchard management.

Traditional plant disease assessment primarily relies on manual field inspections performed by farmers and agricultural experts. Although these approaches provide reliable diagnoses, they are labour-intensive, time-consuming, subjective, and difficult to scale across large agricultural systems. Consequently, considerable research efforts have focused on developing automated disease assessment systems capable of supporting precision agriculture through rapid, objective, and scalable disease monitoring.

Recent advances in artificial intelligence (AI), computer vision, and deep learning have revolutionized automated plant disease diagnosis. Convolutional Neural Networks (CNNs), Vision Transformers (ViTs), and related deep learning architectures have demonstrated remarkable performance for disease classification, segmentation, severity estimation, and symptom recognition across numerous crops \cite{pacal2025,upadhyay2025}. These advances have been driven largely by the increasing availability of annotated plant disease datasets, which serve as the foundation for training and evaluating modern computer vision models. High-quality datasets not only determine the generalization capability of learning algorithms but also establish standardized benchmarks that enable fair comparison among competing methods. Consequently, the availability, quality, diversity, and representativeness of datasets have become critical factors influencing progress in AI-based plant disease assessment \cite{ngugi2024machine,shafik2023systematic}.

The remarkable success of deep learning for plant disease recognition has been largely enabled by the availability of publicly accessible image datasets such as PlantVillage \cite{hughes2015}, PlantDoc \cite{singh2020}, Plant Pathology 2020 \cite{thapa2020}, AI Challenger \cite{liang2019}, and several other crop disease collections. These datasets have established valuable benchmarks for disease classification and have significantly accelerated research in computer vision-based plant health assessment.

Despite these developments, disease progression remains an inherently temporal biological process that unfolds continuously over time. The visual symptoms observed on an infected leaf represent only a single stage within a much longer progression trajectory influenced by environmental conditions, pathogen behaviour, host response, and disease management practices. Continuous monitoring of disease evolution therefore provides valuable information regarding severity accumulation, progression dynamics, symptom development, and temporal variability that cannot be inferred from isolated observations. Such temporal information is essential for disease forecasting, progression modelling, early warning systems, precision disease management, and digital representations of crop health. Consequently, longitudinal disease monitoring has emerged as an important direction for developing next-generation intelligent plant disease assessment systems. To systematically analyze these limitations, representative publicly available plant disease datasets are compared in terms of acquisition environment, temporal characteristics, annotation quality, and disease progression information, as summarized in Table~\ref{tab:dataset_comparison}.

First, the majority of publicly available plant disease datasets are inherently static. Each image represents a single observation acquired at one particular point in time, and individual samples are generally treated as independent observations. While such datasets are well suited for conventional disease classification tasks, they provide no information regarding how disease symptoms evolve, expand, or change throughout the infection cycle. Second, many widely used benchmark datasets have been collected under controlled laboratory or experimental conditions. Images are frequently captured using homogeneous backgrounds, controlled illumination, minimal occlusion, and carefully positioned leaves. Although these acquisition settings simplify computer vision tasks and facilitate model development, they fail to represent the substantial environmental variability encountered in commercial orchards, including changing illumination, background clutter, varying viewpoints, overlapping leaves, weather effects, and natural symptom heterogeneity. Third, existing datasets rarely incorporate repeated observations of the same plant organ throughout disease development. Since images are generally collected from different leaves or different plants, there is no temporal correspondence between observations. As a result, disease progression, severity accumulation, symptom expansion, growth dynamics, and temporal variability cannot be studied using these datasets, which restricts research efforts to develop learning models that can leverage historical disease information. Furthermore, annotation protocols employed by existing datasets are often designed primarily for image classification, providing disease category labels without detailed temporal metadata, disease progression records, or standardized severity measurements. Consequently, important downstream tasks such as disease progression modelling, severity forecasting, temporal representation learning, digital twins, and progression-aware disease intelligence remain largely unexplored due to the absence of suitable benchmark datasets. These limitations highlight a fundamental gap between current benchmark datasets and the practical requirements of real-world disease monitoring. Developing reliable longitudinal datasets that capture repeated observations of individual leaves under natural field conditions is therefore essential for advancing temporal plant disease research. Such datasets enable the investigation of disease progression dynamics and support the development of progression-aware computer vision methods.

To address these limitations, this paper presents \textbf{AppleScab-LT}, a real-field longitudinal dataset specifically designed for temporal disease progression analysis of apple scab. The dataset comprises repeated observations of individually tracked leaves acquired under natural orchard conditions, together with expert-verified annotations, severity quantification, temporal metadata, and quality assurance protocols that preserve the continuity of disease evolution. By enabling the investigation of disease progression dynamics rather than isolated disease instances, AppleScab-LT provides a longitudinal dataset for developing and evaluating next-generation temporal disease intelligence systems.

\subsection{Primary Research Question}
\label{sec:Grand Research Question}
The limitations identified in existing plant disease datasets reveal that although numerous datasets support disease classification and recognition, they largely overlook the temporal nature of plant disease progression under real-world agricultural conditions by treating each image as an independent observation. Consequently, fundamental aspects of disease evolution, including symptom progression, severity accumulation, temporal variability, and repeated monitoring of the same infected tissue, remain inadequately represented. To address these limitations, this study is guided by the following Primary Research Question (PRQ):

\begin{quote}
\textbf{How can a reliable longitudinal dataset be developed for studying plant disease progression under real-field orchard conditions while ensuring annotation quality, temporal consistency, and scientific reliability?}
\end{quote}

Answering this question requires a comprehensive investigation extending beyond simple image acquisition. It involves designing a systematic data collection protocol, establishing reliable annotation and quality assurance procedures, validating temporal consistency, characterizing disease progression dynamics, and defining standardized evaluation protocols for future temporal plant disease research. Accordingly, the remainder of this paper is organized around a series of complementary research questions that collectively address the Primary  Research Question.

\subsection{Research Questions}
\label{sec:Research Questions}
To systematically address the Primary Research Question, the proposed study is structured around five interrelated research questions that collectively describe the complete lifecycle of developing a reliable longitudinal dataset. These research questions guide the methodology, analysis, and evaluation presented throughout the paper and establish a logical progression from identifying existing limitations to developing a temporal disease dataset.

The relationships among these research questions summarise the overall research framework adopted in this study and illustrates how each research question contributes toward answering the Primary Research Question. Beginning with the limitations of existing datasets, the framework proceeds through dataset development, reliability assessment, dataset characterization, and finally the definition of standardized evaluation protocols for future temporal disease studies.

\begin{itemize}

\item \textbf{RQ1: Why are existing plant disease datasets insufficient for studying disease progression?}

This question investigates the limitations of current plant disease datasets with respect to temporal monitoring, repeated observations, disease progression representation, and real-field applicability. Addressing this question establishes the motivation for developing a new longitudinal dataset.

\item \textbf{RQ2: How can a reliable longitudinal apple scab dataset be developed under real-field conditions?}

This question focuses on the systematic development of the proposed dataset, including orchard monitoring, repeated leaf tracking, image acquisition, expert-guided disease verification, and temporal sequence construction.

\item \textbf{RQ3: How can annotation quality and temporal reliability be ensured in a longitudinal disease dataset?}

This question addresses the scientific reliability of the dataset through standardized annotation protocols and guidelines, expert disease validation, quality assurance procedures, temporal assessment and consistency, and reliability validation.

\item \textbf{RQ4: What temporal disease characteristics are captured within the proposed longitudinal dataset?}

This question investigates the statistical and temporal characteristics of the collected data, including disease progression dynamics, inter-leaf variability, severity evolution, growth behaviour, and disease stage distribution.

\item \textbf{RQ5: How can the proposed dataset support future temporal plant disease research?}

This question establishes the long-term research value of the dataset by defining standardized data organization, metadata, evaluation protocols, and potential downstream research tasks including disease progression modelling, severity estimation, forecasting, segmentation, and temporal learning.

\end{itemize}

Rather than presenting the dataset as a collection of images, the framework that presents dataset development as a systematic scientific process consisting of five interconnected stages, where each stage addresses a specific research question and collectively contributes to the development of a reliable longitudinal dataset. The first two research questions establish the motivation and systematic construction of the dataset, whereas the subsequent questions focus on ensuring scientific reliability, understanding the temporal characteristics captured by the dataset, and demonstrating its value as a high-quality resource for the broader research community. This research-question-oriented organization provides a coherent framework that guides the remainder of the paper.

\subsection{Contributions}
\label{sec:Contributions}
The major contributions of this work are summarized as follows:

\begin{enumerate}

\item \textbf{Development of AppleScab-LT: A Longitudinal Real-Field Apple Scab Dataset.}

A novel longitudinal dataset, namely \textbf{AppleScab-LT}, was developed through systematic monitoring of naturally infected apple leaves under real orchard conditions. Unlike conventional static datasets, the proposed dataset captures repeated observations of the same infected leaves throughout disease progression, providing valuable temporal information for studying disease evolution.

\item \textbf{Expert-Guided Annotation and Reliable Dataset Curation Framework.}

A comprehensive dataset curation pipeline was established incorporating expert-guided disease verification, polygon-based annotation, leaf isolation, disease severity quantification, metadata generation, and quality assurance procedures to ensure annotation accuracy and dataset consistency.

\item \textbf{Reliability Assessment for Longitudinal Disease Datasets.}

A dedicated reliability validation framework was introduced to evaluate annotation consistency, temporal sequence integrity, expert verification, and quality assurance throughout the dataset development process, thereby improving the scientific reliability and reproducibility of the proposed dataset.

\item \textbf{Comprehensive Characterization of Disease Progression Dynamics.}

The temporal characteristics of the dataset were systematically analyzed through disease progression dynamics, severity evolution, inter-leaf variability, temporal growth patterns, and stage-wise severity distributions, that help in providing insights into the pattern of apple scab development under natural orchard conditions.

\item \textbf{Longitudinal Dataset for Temporal Plant Disease Research.}

Beyond providing a curated image repository, AppleScab-LT establishes standardized evaluation protocols, metadata organization, and evaluation guidelines to support a wide range of downstream tasks, including disease stage classification, disease severity estimation, progression forecasting, temporal representation learning, digital twins, and other progression-aware plant disease applications.

\end{enumerate}

\paragraph{Paper Organization}
The remainder of this paper is organized according to the five research questions introduced in Section~\ref{sec:Introduction}. Section~\ref{sec:related_work} addresses \textbf{RQ1} by reviewing existing plant disease datasets and identifying the limitations of current datasets for longitudinal disease progression analysis. Section~\ref{sec:dataset_development} answers \textbf{RQ2} by describing the development of AppleScab-LT, study area, acquisition system, longitudinal monitoring protocol, and construction methodology. Section~\ref{sec:annotation} addresses \textbf{RQ3} by presenting the annotation protocol, expert verification process, quality assurance, and reliability assessment framework. Section~\ref{sec:dataset_characterization} answers \textbf{RQ4} through a comprehensive characterization of the dataset, including statistics, temporal disease severity analysis, progression dynamics, growth behaviour, disease stage construction, and dataset diversity. Section~\ref{sec:Dataset Organization and Research Utility} addresses \textbf{RQ5} by introducing the dataset organization, supported research tasks, recommended evaluation protocol, reference baseline models, and reproducibility guidelines. Finally, Sections~\ref{sec:discussion}--\ref{sec:conclusion} present discussion, limitations and future work, and conclusion. 

\section{Related Work}
\label{sec:related_work}
The rapid advancement of artificial intelligence in plant disease diagnosis has been driven largely by the availability of publicly accessible image datasets. Over the past decade, numerous datasets have been introduced to facilitate disease classification, detection, and segmentation across a wide range of crops. These datasets have significantly accelerated the development of computer vision / deep learning models by providing standardized benchmarks for algorithm evaluation. However, despite their importance, most existing datasets have been designed primarily for static image analysis and provide limited support for studying disease progression under natural field conditions.

This section reviews various plant disease datasets with particular emphasis on their acquisition conditions, annotation characteristics, temporal information, and applicability to progression-aware disease analysis. 

\subsection{Plant Disease Image Datasets}
\label{sec:Plant Disease Image Datasets}
Publicly available plant disease image datasets have played a fundamental role in advancing automated plant disease diagnosis using machine learning and deep learning techniques. Among these, the PlantVillage dataset \cite{hughes2015} remains the most influential dataset, containing more than 50,000 images covering multiple crop species and disease categories. Owing to its large-scale crop collection, PlantVillage has been extensively employed for developing disease classification algorithms. Nevertheless, images were acquired under controlled laboratory conditions using homogeneous backgrounds and carefully isolated leaves, limiting their representativeness for practical field deployment. Additional lab-based datasets available for reference are detailed in Table~\ref{tab:dataset_comparison}.

\begin{table*}[hbt!]
\centering
\caption{Comparison of representative plant disease datasets with respect to field acquisition, temporal monitoring, annotation characteristics, and progression information.}
\label{tab:dataset_comparison}
\vspace{-3mm}
\resizebox{\textwidth}{!}{
\begin{tabular}{lcccccccc}
\toprule
Dataset & Crop & Field & Temporal & Longitudinal & Segmentation & Expert Verified & Disease Severity & Repeated Monitoring \\
\midrule
\multicolumn{9}{@{}l}{\textbf{Single-Plant Datasets}} \\
\midrule
ALDD Apple Dataset [Jiang et al., 2019] \cite{jiang2019} & 1 & No & No & No & No & Yes & No & No \\
Soybean Dataset [Wu et al., 2019] \cite{wu2019}  & 1 & No & No & No & No & Yes & No & No \\
Citrus Dataset [Dananjayan et al., 2022] \cite{dananjayan2022} & 1 & No & No & No & No & Yes & No & No \\
Rice Dataset [Joshi et al., 2022] \cite{joshi2022} & 1 & No & No & No & No & Yes & No & No \\
Apple Plant Dataset [Khan et al., 2022] \cite{khan2022} & 1 & No & No & No & No & Yes & No & No \\
Tomato Dataset [Chug et al., 2023]\cite{chug2023} & 1 & Yes & No & No & No & Yes & No & No \\
Tomato Dataset [Khatoon et al., 2021] \cite{khatoon2021} & 1 & Yes & No & No & No & Yes & No & No \\
Tomato Dataset [Ashwinkumar et al., 2022] \cite{ashwinkumar2022} & 1 & No & No & No & No & Yes & No & No \\
Rice Dataset [Prajapati et al., 2017]\cite{prajapati2017} & 1 & No & No & No & No & Yes & No & No \\
Soybean Dataset [Bevers et al., 2022] \cite{bevers2022} & 1 & No & No & No & No & Yes & No & No \\
Citrus Dataset [Syed et al., 2022]\cite{syed2022} & 1 & No & No & No & No & Yes & No & No \\
Coffee Dataset [Tassis et al., 2021] \cite{tassis2021} & 1 & Yes & No & No & No & Yes & No & No \\
Sugar Dataset [Adem et al., 2023] \cite{adem2023} & 1 & Yes & No & No & No & Yes & No & No \\
Beans Dataset [Elfatimi et al., 2022] \cite{elfatimi2022} & 1 & No & No & No & No & Yes & No & No \\
Citrus Dataset [Liu et al., 2022] \cite{liu2022} & 1 & No & No & No & No & Yes & No & No \\
Banana Dataset [Seetharaman et al., 2022] \cite{seetharaman2022} & 1 & Yes & No & No & No & Yes & No & No \\
Grape Dataset [Thet et al., 2020] \cite{thet2020} & 1 & No & No & No & No & Yes & No & No \\
Wheat Dataset [Rangarajan et al., 2022] \cite{rangarajan2022} & 1 & No & No & No & No & Yes & No & No \\
Grape Dataset [Perez et al., 2022]\cite{perez2022} & 1 & No & No & No & No & Yes & No & No \\
Plant Pathology 2020 [Thapa et al.] \cite{thapa2020} & 1 & Yes & No & No & No & Yes & No & No \\
Arabica coffee Dataset [Jepkoech et al.] \cite{jepkoech2021} & 1 & Yes & No & No & No & Yes & No & No \\
iCassava 2019 [Mwebaze et al., 2019] \cite{mwebaze2019} & 1 & Yes & No & No & No & Partial & No & No \\
Makerere Bean Disease [Makerere AI Lab, 2020]\cite{ibean_dataset} & 1 & Yes & No & No & No & Yes & No & No \\
GLDD [Xie et al., 2020]\cite{xie2020} & 1 & No & No & No & No & Yes & No & No \\
LDSD [Fakhre et al., 2021] \cite{fakhre2021} & 1 & Yes & No & No & Yes & Yes & No & No \\
NLB Dataset [Prashanth et al., 2024] \cite{prashanth2024} & 1 & Yes & No & No & Yes & Yes & Yes & No \\
NLB Maize Dataset [DeChant et al., 2017] \cite{dechant2017} & 1 & Yes & No & No & Yes & Yes & Partial & No \\
ADLD (Apple Leaf Disease) [Yulong et al. 2026] \cite{fan2026} & 1 & Yes & No & No & Yes & Yes & No & No \\
Pigeon Pea Disease [Bhagat et al., 2024] \cite{bhagat2024} & 1 & Yes & No & No & No & Yes & No & No \\
\midrule
\multicolumn{9}{@{}l}{\textbf{Multi-Plant Datasets}} \\
\midrule
PlantVillage Dataset [Hughes et al., 2015] \cite{hughes2015} & 14 & No & No & No & No & Yes & No & No \\
PlantDoc Dataset [Singh et al., 2020] \cite{singh2020} & 13 & Partial & No & No & No & Yes & No & No \\
NZDL Fruit Dataset [Saleem et al., 2022] \cite{saleem2022} & 5 & Yes & No & No & No & Yes & No & No \\
AI Challenger Dataset [Liang et al., 2019] \cite{liang2019} & 10 & No & No & No & No & Yes & No & No \\
Turkey Plant Dataset [Tugrul et al., 2022] \cite{tugrul2022} & 7 & Yes & No & No & No & Yes & No & No \\
Multi Plant Dataset [Chouhan et al., 2019] \cite{chouhan2019} & 11 & No & No & No & No & Yes & No & No \\
FieldPlant Dataset [Moupojou et al., 2023] \cite{moupojou2023} & 3 & Yes & No & No & No & Yes & No & No \\
Plant Disease Dataset-271 (PDD271) [Liu et al., 2021] \cite{liu2021} & 38 & Yes & No & No & No & Yes & No & No \\
UCI-Rice and Wheat Dataset [Jiang et al., 2021] \cite{jiang2021uci} & 2 & No & No & No & No & Yes & No & No \\
PlantDet Dataset (Rice and Betel) [Shovon et al., 2023] \cite{shovon2023} & 2 & Partial & No & No & No & Yes & No & No \\
Multispectral Image Dataset [Brown et al., 2023] \cite{brown2024} & 5 & Yes & No & No & No & Yes & No & No \\
PlantWild [Wei et al., 2024] \cite{wei2024} & 33 & Yes & No & No & No & Yes & No & No \\
PlantSeg [Wei et al., 2024] \cite{wei2024a} & 34 & Yes & No & No & Yes & Yes & No & No \\
LeafNet [Quoc et al., 2025] \cite{quoc2026} & 22 & No & No & No & No & Yes & No & No \\
MuST-C [Chong et al., 2026] \cite{chong2026} & 6 & Yes & Yes & Yes & No & Yes & No & Yes \\
MSMSVDD [Li et al., 2022] \cite{li2022} & 5 & Yes & No & No & No & Yes & No & No \\

\bottomrule
\end{tabular}}
\end{table*}

To overcome the limitations of laboratory datasets, several field-acquired datasets have subsequently been introduced, as detailed in Table~\ref{tab:dataset_comparison}. Some of them are the Fieldlant dataset \cite{moupojou2023}, which incorporates images collected under natural agricultural environments, thereby introducing realistic challenges such as varying illumination, complex backgrounds, partial occlusions, and multiple viewing angles. Similarly, the Plant Pathology 2020 dataset \cite{thapa2020} provided field images of apple leaves captured under orchard conditions. Other datasets, including PlantSeg dataset \cite{wei2024a} and Arabica coffee dataset \cite{jepkoech2021}, further expanded the availability of field-acquired plant disease datasets.

Despite their diversity, these datasets primarily support conventional image classification and disease recognition tasks. These datasets cannot be used to study  disease evolution, severity progression, symptom expansion, or temporal growth dynamics because individual images are typically treated as independent observations with no temporal relationship between successive samples. Furthermore, detailed longitudinal metadata, repeated observations of the same infected tissue, and progression histories are generally absent. Although existing datasets have substantially advanced computer vision research in agriculture, none simultaneously provide repeated monitoring of individual leaves, longitudinal progression records, standardized disease severity measurements, comprehensive segmentation annotations, and temporal benchmark protocols.

\subsection{Apple Disease Datasets}
\label{sec:}
Apple production has received considerable attention within the plant disease research community owing to the significant economic impact of diseases such as apple scab, cedar apple rust, black rot, powdery mildew, and Marssonina leaf blotch. Consequently, several publicly available datasets have been developed specifically for apple disease recognition. Among the earliest and most widely used resources, the PlantVillage dataset \cite{hughes2015} contains laboratory-acquired images of healthy and diseased apple leaves. Although the dataset has served as the foundation for various deep learning studies, its controlled acquisition environment does not adequately represent the visual complexity encountered in commercial orchards. To bridge the gap, several real-world datasets have been developed, such as ADLD (Apple Leaf Disease) \cite{fan2026}, which introduced field-acquired apple leafs containing images captured directly within orchards. These datasets incorporate variations in illumination, background clutter, leaf orientation, and symptom appearance, thereby providing more realistic benchmarks for disease classification. Similarly, several researchers have developed smaller field datasets focusing on apple scab, powdery mildew, or multiple apple diseases under natural orchard conditions. Table~\ref{tab:apple_dataset_comparison} compares several apple disease datasets. 

Despite these advances, existing apple disease datasets have several limitations. As images are generally acquired as isolated observations without preserving temporal relationships among successive monitoring events. Repeated observations of the same infected leaf are rarely available, preventing the study of disease progression trajectories. Disease severity is often represented only through categorical disease labels rather than quantitative progression measurements. Also, metadata describing monitoring intervals, temporal sequence identity, or disease evolution are generally unavailable. This limitation significantly restricts research on various progression-aware applications.

\begin{table*}[!htbp]
\centering
\caption{Comparison of representative apple disease datasets.}
\label{tab:apple_dataset_comparison}
\vspace{-3mm}
\resizebox{\textwidth}{!}{
\begin{tabular}{lccccccc}
\toprule
Dataset & Field & Longitudinal & Repeated Leaf Tracking & Severity Labels & Segmentation & Temporal Metadata & Progression Analysis \\
\midrule
\multicolumn{8}{@{}l}{\textbf{Single-Plant Apple Datasets}} \\
\midrule
ALDD Apple Dataset \cite{jiang2019} & $\times$ & $\times$ & $\times$ & $\times$ & $\times$ & $\times$ & $\times$ \\
Apple Plant Dataset \cite{khan2022} & $\times$ & $\times$ & $\times$ & $\times$ & $\times$ & $\times$ & $\times$ \\
Plant Pathology Dataset \cite{thapa2020} & \checkmark & $\times$ & $\times$ & $\times$ & $\times$ & $\times$ & $\times$ \\
ADLD (Apple Leaf Disease) \cite{fan2026} & \checkmark & $\times$ & $\times$ & $\times$ & \checkmark & $\times$ & $\times$ \\
\midrule
\multicolumn{8}{@{}l}{\textbf{Multi-Plant Datasets (containing Apple)}} \\
\midrule
PlantVillage Dataset \cite{hughes2015} & $\times$ & $\times$ & $\times$ & $\times$ & $\times$ & $\times$ & $\times$ \\
PDD Dataset-271 (PDD271) \cite{liu2021} & \checkmark & $\times$ & $\times$ & $\times$ & $\times$ & $\times$ & $\times$ \\ 
PlantDoc Dataset \cite{singh2020} & $\circleddash$  & $\times$ & $\times$ & $\times$ & $\times$ & $\times$ & $\times$ \\
Turkey Plant Dataset \cite{tugrul2022} & \checkmark  & $\times$ & $\times$ & $\times$ & $\times$ & $\times$ & $\times$ \\
NZDL Fruit Dataset \cite{saleem2022} & \checkmark  & $\times$ & $\times$ & $\times$ & $\times$ & $\times$ & $\times$ \\
AI Challenger Dataset \cite{liang2019} & $\times$  & $\times$ & $\times$ & $\times$ & $\times$ & $\times$ & $\times$ \\
PlantWild Dataset \cite{wei2024} & \checkmark  & $\times$ & $\times$ & $\times$ & $\times$ & $\times$ & $\times$ \\
PlantSeg Dataset \cite{wei2024a} & \checkmark & $\times$ & $\times$ & $\times$ & \checkmark & $\times$ & $\times$ \\
LeafNet Dataset \cite{quoc2026} & $\times$ & $\times$ & $\times$ & $\times$ & $\times$ & $\times$ & $\times$ \\
\midrule
\rowcolor[gray]{0.92}
AppleScab-LT (Proposed) & \checkmark & \checkmark & \checkmark & \checkmark & \checkmark & \checkmark & \checkmark \\
\bottomrule
\end{tabular}
}
\begin{minipage}{17cm}
\footnotesize\itshape{\textbf{Symbol meanings:} $\checkmark$ = Available; $\circleddash$ = Partially Available; $\times$ = Not Available.}
\end{minipage}
\end{table*}

\subsection{Longitudinal Agricultural Datasets}
\label{sec:Longitudinal Agricultural Datasets}
Longitudinal datasets have become increasingly important in agricultural research owing to their ability to capture temporal variations that cannot be observed using isolated measurements. Unlike conventional image datasets, longitudinal datasets preserve the temporal continuity of biological processes by repeatedly monitoring the same object or experimental unit over time. Such datasets support numerous applications, including crop growth modelling, phenological analysis, yield prediction, environmental monitoring, and precision agriculture.

Longitudinal datasets specifically designed for plant disease monitoring remain extremely scarce. Some agricultural studies have collected temporal data using remote sensing platforms, unmanned aerial vehicles (UAVs), fixed camera systems, and environmental sensor networks. These datasets generally record crop development to investigate seasonal growth patterns, biomass accumulation, canopy development, and vegetation dynamics \cite{feng2025,chong2026, dang2026}. Although these studies demonstrate the value of temporal monitoring, they differ fundamentally from the requirements of disease progression analysis. Most available longitudinal agricultural datasets focus primarily on crop growth rather than disease development. Furthermore, many datasets record temporal measurements at the field or canopy level without preserving repeated observations of individual leaves or disease lesions. Consequently, providing limited support for understanding the temporal evolution of plant diseases at the organ level.

A systematic comparison of available representative general and longitudinal plant disease datasets is summarised in Table~\ref{tab:dataset_taxonomy}. The datasets are grouped into two categories, namely general plant disease datasets and longitudinal plant disease monitoring dataset. The comparison evaluates acquisition environment, repeated monitoring capability, longitudinal observations, temporal severity annotations, manual polygon annotation, metadata and temporal descriptors, expert disease verification, leaf-level tracking, and disease progression analysis.

\begin{table*}[!htbp]
\centering
\caption{
Taxonomy and comparison of representative general and longitudinal plant disease datasets.}
\vspace{-3mm}
\label{tab:dataset_taxonomy}

\renewcommand{\arraystretch}{1.3}

\resizebox{\textwidth}{!}{
\begin{tabular}{llccccccccccc}

\hline

\textbf{Category}
&
\textbf{Dataset}
&
\textbf{Crop}
&
\textbf{Field}
&
\textbf{Repeated}
&
\textbf{Longitudinal}
&
\textbf{Temporal}
&
\textbf{Manual}
&
\textbf{Metadata \&}
&
\textbf{Expert}
&
\textbf{Leaf-level}
&
\textbf{Disease}
\\

&&&
&
\textbf{Monitoring}
&
&
\textbf{Severity Labels}
&
\textbf{Polygon Annotation}
&
\textbf{Temporal Descriptors}
&
\textbf{Disease Verification}
&
\textbf{Tracking}
&
\textbf{Progression Analysis}
\\

\hline

\multirow{10}{*}{\makecell{General Plant \\ Datasets}}

& PlantVillage \cite{hughes2015}
& Multiple
& $\times$
& $\times$
& $\times$
& $\times$
& $\times$
& $\times$
& $\checkmark$
& $\times$
& $\times$
\\

& PlantDoc \cite{singh2020}
& Multiple
& $\checkmark$
& $\times$
& $\times$
& $\times$
& $\times$
& $\times$
& $\checkmark$
& $\times$
& $\times$
\\

& PDD271 \cite{liu2021}
& Multiple
& $\checkmark$
& $\times$
& $\times$
& $\times$
& $\times$
& $\times$
& $\checkmark$
& $\times$
& $\times$
\\

& PlantSeg \cite{wei2024a}
& Multiple
& $\checkmark$
& $\times$
& $\times$
& $\times$
& $\checkmark$
& $\times$
& $\checkmark$
& $\times$
& $\times$
\\

& ADLD \cite{fan2026}
& Apple
& $\checkmark$
& $\times$
& $\times$
& $\times$
& $\checkmark$
& $\times$
& $\checkmark$
& $\times$
& $\times$
\\

& Plant Pathology 2020 \cite{thapa2020}
& Apple
& $\checkmark$
& $\times$
& $\times$
& $\times$
& $\times$
& $\times$
& $\checkmark$
& $\times$
& $\times$
\\

& NLB \cite{prashanth2024}
& Maize
& $\checkmark$
& $\times$
& $\times$
& $\times$
& $\checkmark$
& $\times$
& $\checkmark$
& $\times$
& $\times$
\\

& Pigeon Pea Disease \cite{bhagat2024}
& Maize
& $\checkmark$
& $\times$
& $\times$
& $\times$
& $\times$
& $\times$
& $\checkmark$
& $\times$
& $\times$
\\

\hline

\multirow{4}{*}{\makecell{Longitudinal Plant \\ Datasets}}

& SYMPATHIQUE \cite{anderegg2024}
& Wheat
& $\checkmark$
& $\checkmark$
& $\checkmark$
& $\times$
& $\times$
& $\circleddash$
& $\times$
& $\checkmark$
& $\circleddash$
\\

& UAV Wheat Phenology  \cite{feng2025}
& Wheat
& $\checkmark$
& $\checkmark$
& $\checkmark$
& $\times$
& $\times$
& $\times$
& $\checkmark$
& $\times$
& $\times$
\\

& MuST-C \cite{chong2026}
& Multiple
& $\checkmark$
& $\checkmark$
& $\checkmark$
& $\checkmark$
& $\times$
& $\times$
& $\checkmark$
& $\times$
& $\times$
\\

& CropMap \cite{dang2026} 
& Radish, Cabbage 
& $\checkmark$ 
& $\checkmark$ 
& $\checkmark$ 
& $\times$ 
& $\circleddash$ 
& $\times$
& $\checkmark$ 
& $\times$ 
& $\times$ \\

\hline

\rowcolor[gray]{0.92}

\textbf{Proposed}

&
\textbf{AppleScab-LT}

&
\textbf{Apple}

&
$\checkmark$

&
$\checkmark$

&
$\checkmark$

&
$\checkmark$

&
$\checkmark$

&
$\checkmark$

&
$\checkmark$

&
$\checkmark$

&
$\checkmark$

\\

\hline

\end{tabular}
}
\begin{minipage}{17cm}
\footnotesize\itshape{ \textbf{Symbol meanings:} $\checkmark$ = Available; $\circleddash$ = Partially Available; $\times$ = Not Available.}
\end{minipage}
\end{table*}

As shown in Table~\ref{tab:dataset_taxonomy}, conventional plant disease datasets predominantly comprise independent static observations and generally lack repeated monitoring of the same infected leaves, longitudinal disease progression information, temporal severity annotations, and structured temporal descriptors required for modelling disease evolution. Although several datasets provide segmentation masks or disease severity labels, they remain primarily designed for image classification or segmentation tasks rather than progression-aware disease analysis.

Recent longitudinal plant disease monitoring datasets have introduced repeated observations of individual leaves and temporal image sequences under field conditions, thereby facilitating disease monitoring over time. Nevertheless, these datasets primarily focus on image registration, lesion tracking, or phenotyping applications and generally do not provide temporal severity labels, manually annotated polygon-based lesion boundaries, comprehensive temporal descriptors, expert disease verification protocols, or systematic disease progression characterization.

AppleScab-LT bridges these research domains by integrating repeated monitoring of individual infected leaves under real-field orchard conditions with expert-guided manual polygon annotation, quantitative temporal severity measurements, sequence-level temporal descriptors, comprehensive reliability assessment, and detailed disease progression analysis within a unified longitudinal dataset framework. Collectively, these characteristics distinguish AppleScab-LT from existing plant disease datasets and establish a standardized longitudinal dataset for temporal disease intelligence and progression-aware research.

\subsection{Research Gap Analysis}
\label{sec:Research Gap Analysis}
The preceding review demonstrates that considerable progress has been achieved in developing image datasets for plant disease recognition and longitudinal datasets for general agricultural monitoring. Nevertheless, a critical gap remains between existing benchmark resources and the practical requirements of progression-aware plant disease analysis. Current general plant disease datasets primarily emphasize disease classification using static images acquired at a single observation time. Although several datasets have introduced realistic field conditions, they continue to treat disease assessment as an image recognition problem without preserving temporal continuity between successive observations. Conversely, existing longitudinal agricultural datasets focus predominantly on crop growth, phenotyping, environmental monitoring, and yield estimation rather than disease progression. While these datasets incorporate temporal observations, they generally lack repeated monitoring of individual infected leaves, expert-guided disease annotations, lesion segmentation, quantitative severity measurements, and temporal disease metadata necessary for progression-aware disease modelling. Collectively, the literature reveals several important research gaps:

\begin{enumerate}
\setlength{\itemsep}{0pt}
\setlength{\parsep}{0pt}

\item The absence of publicly available longitudinal datasets specifically designed for monitoring plant disease progression under real-field conditions.

\item The lack of repeated observations of the same infected leaf throughout the disease lifecycle.

\item The limited availability of standardized disease severity measurements that are advantageous for uncovering temporal progression analysis.

\item The absence of comprehensive datasets that can integrate image data, annotation files, temporal metadata, and disease progression information within a combined framework.

\item The lack of standardized evaluation  protocols supporting progression-aware research tasks such as disease forecasting, temporal severity estimation, disease progression modelling, and digital twin development.

\end{enumerate}

These limitations motivate for the development of the proposed \textbf{AppleScab-LT} dataset. Unlike existing  datasets, AppleScab-LT was constructed to observe disease progression through continuous monitoring of apple leaves infected by disease under real field conditions. With the process of integrating longitudinal image sequences and other metadata, the proposed dataset establishes a comprehensive resource for future research on temporal plant disease.

\begin{tcolorbox}[colback=blue!5!white,
                  colframe=blue!60!black,
                  title=\textbf{Answer to RQ1},
                  fonttitle=\bfseries,
                  fontupper=\footnotesize]

The analysis of existing plant disease datasets demonstrates that current datasets have primarily been designed for static disease recognition rather than understanding disease development over time. Although these datasets have significantly advanced image-based classification, segmentation, and detection tasks, most lack repeated monitoring of the same plant organs, temporal continuity, disease evolution information, quantitative progression measurements, and standardized protocols required for longitudinal disease modelling. The comparative evaluation of various plant disease datasets identifies a clear gap between available resources and the requirements of progression-aware artificial intelligence systems. These findings establish the need for a dedicated longitudinal dataset capable of preserving real-field disease evolution through repeated observations. AppleScab-LT was therefore developed to address this gap and provide a longitudinal dataset for temporal plant disease progression research.

\end{tcolorbox}

\section{RQ2: Dataset Development}
\label{sec:dataset_development}
To address the second research question, namely, \emph{``How can a reliable longitudinal dataset be developed under real-field conditions?''}, a systematic dataset development framework was established that combines real-world orchard monitoring, repeated leaf tracking, expert disease verification, standardized image acquisition, and temporal sequence construction. Unlike conventional plant disease datasets that consist of independent observations, the proposed framework was specifically designed to preserve the temporal continuity of disease progression by repeatedly monitoring the same infected leaves throughout the infection cycle.

The overall dataset development methodology consists of five major stages: (i) selection of representative orchard locations, (ii) longitudinal monitoring of naturally infected apple leaves, (iii) standardized image acquisition under real-field conditions, (iv) temporal organization of repeated observations into individual leaf sequences, and (v) generation of a curated longitudinal dataset. 

\subsection{Study Area}
\label{sec:Study Area}
\begin{figure*}[!htbp]
\centering
\includegraphics[width=0.45\textwidth]{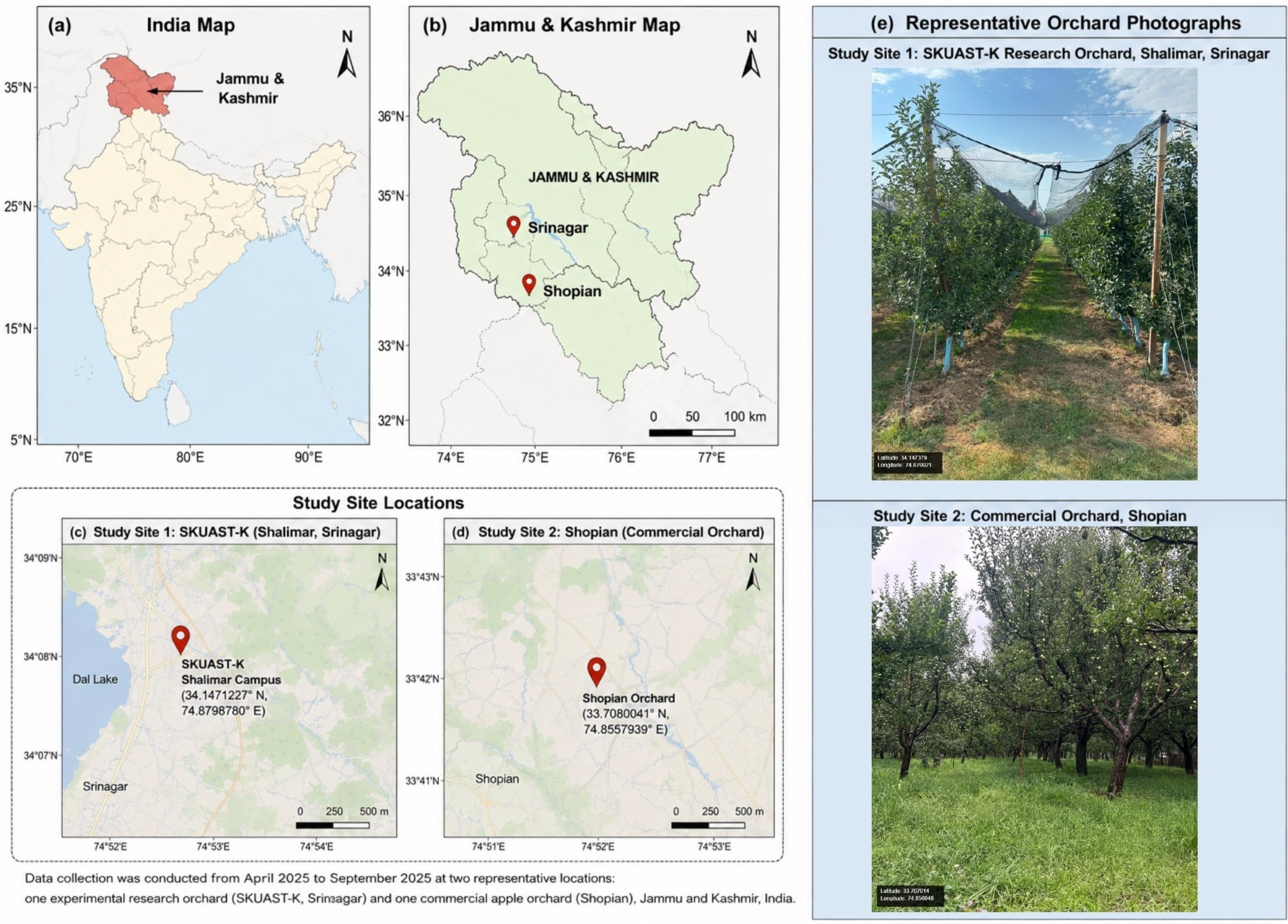}
\vspace{-3mm}
\caption{Geographical locations of the two study locations used for constructing the AppleScab-LT dataset.}
\label{fig:study_area}
\end{figure*}

The development of a reliable longitudinal disease progression dataset requires repeated observations under representative orchard conditions that capture the natural variability of disease development. Accordingly, data collection was conducted from \textbf{April 2025 to September 2025}, corresponding to the complete apple growing season in the Kashmir valley, during which apple scab symptoms emerge, expand, and reach advanced stages of infection. This monitoring period enabled continuous observation of disease progression. To ensure the diversity and representativeness of the collected data, image acquisition was performed at two geographically distinct locations within the Kashmir region, India, as illustrated in Fig.~\ref{fig:study_area}.

These locations were selected because of their importance in apple cultivation and their contrasting orchard environments. The first study site was the \textbf{Sher-e-Kashmir University of Agricultural Sciences and Technology of Kashmir (SKUAST-K)}, located at Shalimar, Srinagar, Jammu and Kashmir, India. The university campus maintains experimental apple orchards that provide a controlled yet naturally occurring environment for monitoring disease progression.\hide{The availability of horticultural experts and research infrastructure facilitated continuous monitoring, expert disease verification, and repeated observations throughout the growing season.} The second study site was a commercial apple orchard located in the \textbf{Shopian district} of Jammu and Kashmir, India, which is widely known owing to its extensive apple cultivation and significant contribution to the horticultural economy of the region. The commercial orchard environment provided naturally occurring disease incidence under practical farming conditions, thereby introducing realistic variability in illumination, canopy structure, orchard management practices, and environmental conditions. Both locations experience a temperate climatic regime characterized by cold winters, mild summers, and seasonal rainfall, creating highly favorable conditions for the development of apple scab caused by \textit{Venturia inaequalis}. Data collection focused exclusively on naturally infected leaves without any artificial inoculation or controlled laboratory intervention, thereby preserving the authentic temporal progression of disease symptoms under real orchard conditions. The environmental and horticultural characteristics of each site are summarized in Table~\ref{tab:study_sites}.

\begin{table*}[!htbp]
\centering
\caption{Characteristics of the study sites selected for longitudinal data collection.}
\label{tab:study_sites}\vspace{-3mm}
\small
\renewcommand{\arraystretch}{0.9}
\begin{tabular}{p{2.7cm}p{6.0cm}p{6.0cm}}
\hline
\textbf{Characteristic} &
\textbf{Study Site 1} &
\textbf{Study Site 2} \\
\hline

Location &
Sher-e-Kashmir University of Agricultural Sciences and Technology of Kashmir (SKUAST-K), Shalimar, Srinagar &
Commercial Apple Orchard, Shopian District \\

Type &
Research Orchard &
Commercial Orchard \\

Region &
Srinagar, Jammu and Kashmir, India &
Shopian, Jammu and Kashmir, India \\

Monitoring Period &
April 2025 -- September 2025 &
April 2025 -- September 2025 \\

Climate &
Temperate Himalayan Climate &
Temperate Himalayan Climate \\

Disease &
Apple Scab (\textit{Venturia inaequalis}) &
Apple Scab (\textit{Venturia inaequalis}) \\

Monitoring Type &
Repeated Leaf Monitoring &
Repeated Leaf Monitoring \\

Environment &
Experimental Research Orchard &
Natural Commercial Orchard \\

Purpose &
Expert-supervised longitudinal monitoring &
Real-field disease progression monitoring \\

\hline
\end{tabular}
\end{table*}

The use of both an experimental research orchard and a commercial production orchard enhances the diversity, ecological validity, and practical applicability of the proposed AppleScab-LT dataset. The two study locations complement each other in terms of orchard characteristics and management practices. The experimental orchard at SKUAST-K enabled systematic monitoring and expert supervision, whereas the commercial orchard in Shopian captured naturally occurring disease progression under practical cultivation conditions. The combination of these two environments improves the diversity and representativeness of the proposed longitudinal dataset. Also, the selected locations differ in terms of orchard type and management practices while sharing climatic conditions favorable for apple scab development. Consequently, AppleScab-LT reflects the variability encountered in practical orchard environments rather than being restricted to a single experimental setting.

\subsection{Data Acquisition System}
\label{sec:Data Acquisition System}
Following the identification of representative orchard sites, a standardized image acquisition system was established to ensure consistent collection of high-quality field images throughout the monitoring campaign. Image acquisition was performed using a Canon EOS R50 mirrorless digital camera equipped with a Canon RF 24--105 mm F4--7.1 IS STM telephoto zoom lens, enabling flexible image acquisition at varying distances while preserving sufficient spatial resolution for subsequent manual annotation and disease severity quantification. The principal specifications of the imaging system are summarized in Table~\ref{tab:camera_specs}.

\begin{table}[!htbp]
\centering
\caption{Specifications of the image acquisition system used during field monitoring.}
\label{tab:camera_specs}
\vspace{-3mm}
\small
\begin{tabular}{ll}
\hline
\textbf{Parameter} & \textbf{Specification} \\
\hline
Camera & Canon EOS R50 \\
Sensor Type & APS-C CMOS \\
Image Resolution & 24.2 Megapixels \\
Lens & Canon RF 24--105 mm F4--7.1 IS STM \\
Sensor Resolution & 24.2 MP APS-C CMOS \\
Image Format & RGB (JPEG) \\
Acquisition Distance & Approximately 0.4--1.0 m \\
Lighting Condition & Natural daylight (field conditions) \\
Image Frequency & Repeated field visits throughout the season \\
\hline
\end{tabular}
\end{table}

To minimize disturbance to the leaves while maintaining consistent image quality, photographs were acquired at an approximate distance of 0.4--1.0~m from the target leaf, depending on canopy accessibility and branch orientation. Images were captured under natural daylight conditions without artificial illumination in order to preserve realistic variability encountered during practical orchard monitoring. Consequently, the dataset includes variations in illumination intensity, viewing angle, shadows, background clutter, partial occlusion, and disease severity, thereby reflecting real operational conditions encountered in commercial apple orchards. Field visits were conducted regularly throughout the growing season from April to September 2025. During each visit, both newly infected leaves and previously monitored leaves were photographed whenever accessible. Multiple images were often acquired from slightly different viewpoints to ensure adequate visual coverage while maintaining image sharpness and focus.

\subsection{Longitudinal Monitoring Protocol}
\label{sec:Longitudinal Monitoring Protocol}

\begin{figure}[!htpb]
\centering
\includegraphics[width=0.3\linewidth]{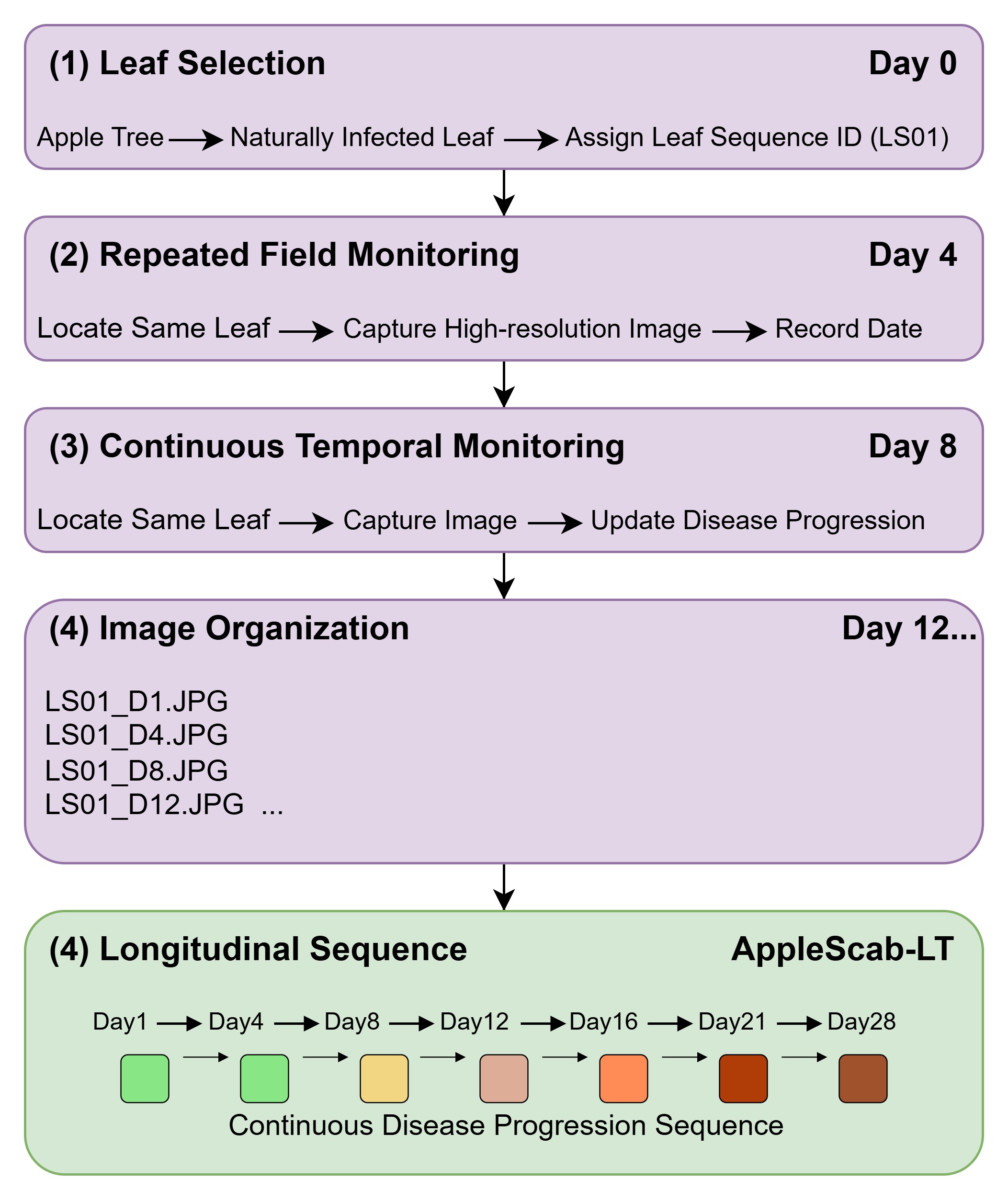}
\vspace{-3mm}
\caption{Longitudinal monitoring protocol adopted for temporal dataset construction.}
\label{fig:longitudinal_protocol}
\end{figure}

It is important to note that the primary objective of the field survey was not limited to apple scab but rather the development of a comprehensive real-world multi-disease apple image repository. Consequently, images representing multiple foliar diseases, insect damage, physiological disorders, and healthy leaves were collected under diverse environmental conditions. The longitudinal apple scab dataset presented in this work was subsequently extracted from this larger multi-disease repository after expert verification and temporal sequence identification. Following the construction of the multi-disease repository, repeated observations recorded during successive field visits revealed that several infected leaves had been photographed multiple times throughout the growing season. These repeated observations enabled the development of a dedicated longitudinal monitoring protocol for studying disease progression. The monitoring strategy focused on identifying the same infected leaves throughout the disease development cycle, and individual leaves were treated as continuous temporal entities whose visual symptoms evolved over time. This strategy enabled the construction of disease progression sequences suitable for temporal analysis. Figure~\ref{fig:longitudinal_protocol} depicts the longitudinal monitoring protocol adopted for temporal dataset construction.

\subsubsection{Leaf Selection}
\label{sec:}
Leaves exhibiting clearly visible disease symptoms and remaining physically intact throughout the monitoring period were selected for longitudinal observation. Priority was given to leaves that could be reliably relocated and whose disease progression could be monitored without significant occlusion or physical damage.

\subsubsection{Leaf Tracking}
\label{sec:}
Selected leaves were chosen that were revisited during successive monitoring sessions using their relative position within the tree canopy, branch architecture, and neighbouring leaves as visual references. This procedure ensured that temporal changes reflected genuine disease progression rather than variations by inconsistent image acquisition.

\vspace{-2mm}
\subsubsection{Monitoring Interval}
\label{sec:}

Since the primary objective was to develop a multi-disease apple disease repository, the monitoring schedule was not designed to follow a fixed observation interval for individual leaves. Instead, field visits were conducted periodically throughout the growing season based on disease occurrence, orchard accessibility, prevailing weather conditions, and practical constraints associated with large-scale data collection. As a result, the temporal intervals between consecutive observations are naturally irregular, reflecting realistic field monitoring practices while preserving authentic disease progression under real-world environmental conditions.

\subsubsection{Image Naming Convention}
\label{sec:}
To maintain chronological organization of the collected data, every image was assigned a unique filename incorporating the monitored leaf identifier, observation day, and image index. The naming convention followed the format

\[
\texttt{LSXX\_DYY\_(Z)}
\]

where \texttt{LSXX} denotes the leaf sequence identifier, \texttt{DYY} represents the monitoring day relative to the first observation, and \texttt{(Z)} indicates multiple images acquired during the same monitoring session when necessary. For example,

\begin{center}
\texttt{LS03\_D15\_(2).JPG}
\end{center}

represents the second image acquired from leaf sequence LS03 on the fifteenth monitoring day.

\subsubsection{Sequence Identification}
\label{sec:}
Each monitored leaf was assigned a unique sequence identifier (LS01--LS20). All images corresponding to the same physical leaf were subsequently organized chronologically according to monitoring day, thereby forming an ordered disease progression sequence. Each sequence represents repeated observations of a single infected leaf throughout its disease development and serves as the fundamental temporal unit of the proposed longitudinal dataset.
The adopted monitoring protocol transforms conventional independent disease images into temporally ordered disease progression records enabling progression-aware disease modeling.

\subsection{Multi-Disease Source Repository}
\label{sec:Multi-Disease Repository}
The field campaign was originally initiated with the objective of developing a comprehensive real-world image repository encompassing multiple diseases affecting apple orchards under natural environmental conditions. The intention was to capture the diversity of foliar diseases encountered by growers during routine orchard management. Consequently, repeated field visits were conducted throughout the growing season to document naturally occurring disease symptoms exhibiting substantial variability in disease severity, symptom morphology, illumination, background complexity, and environmental conditions. During each field visit, symptomatic leaves corresponding to different diseases were photographed using the standardized acquisition protocol presented in Section~\ref{sec:Data Acquisition System}. 

\begin{figure*}[!htbp]
\centering
\includegraphics[width=0.38\textwidth]{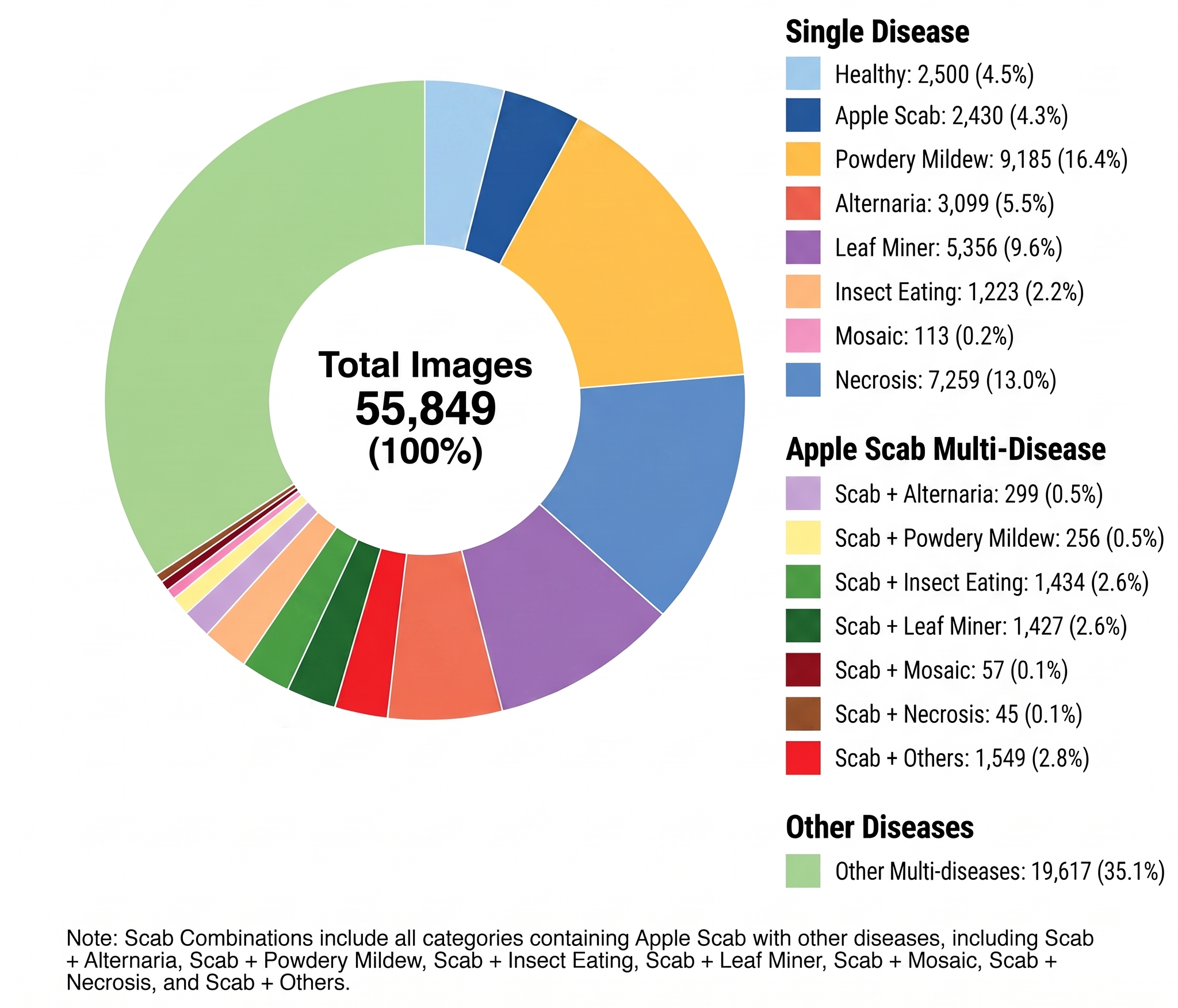}
\vspace{-3mm}
\caption{Overview of the multi-disease apple repository developed during the field campaign.}
\label{fig:multidisease_repository}
\end{figure*}

The resulting repository contains images representing a broad spectrum of foliar diseases and physiological disorders commonly observed in apple production systems. These include apple scab, powdery mildew, Alternaria leaf blotch, mosaic virus, leaf curl, insect-induced damage, leaf miner infestation, nutrient stress symptoms, and healthy leaves. The inclusion of multiple disease categories introduces considerable diversity in symptom appearance, lesion morphology, colour variation, disease severity, and background complexity, making the repository suitable for future studies on disease classification, detection, segmentation, and severity estimation. The disease-wise composition of the multi-disease repository is presented in Figure~\ref{fig:multidisease_repository}. Although the complete repository was originally developed to support multi-class disease recognition, the present work focuses specifically on the longitudinal apple scab subset because of its unique capability to capture disease progression through repeated observations of individual leaves.

During the continuous monitoring campaign, it was observed that several apple scab infected leaves remained attached to the same branches over extended periods while exhibiting progressive symptom development. Unlike other diseases, apple scab lesions gradually expanded over successive observations, providing an opportunity to repeatedly image the same infected leaf throughout the growing season. This observation motivated the extraction of temporal image sequences from the larger multi-disease repository and ultimately led to the development of the AppleScab-LT longitudinal dataset described in the following subsection. Apple scab represents the only disease category that was systematically monitored through repeated observations of individual leaves, thereby enabling the construction of longitudinal disease progression sequences.

\subsection{Construction of AppleScab-LT Dataset}
\label{sec:Identification and Construction AppleScab-LT Dataset}
During the organization and verification of the multi-disease repository, it was observed that a subset of apple scab images corresponded to repeated observations of the same infected leaves acquired over multiple field visits. This unexpected characteristic presented an opportunity to construct a longitudinal disease progression dataset. Consequently, a dedicated curation process was initiated to identify, verify, and organize these repeated observations into temporal leaf sequences, leading to the development of the AppleScab-LT longitudinal dataset. Rather than collecting additional images, the dataset was derived through systematic identification, verification, and organization of repeated observations corresponding exclusively to apple scab infected leaves.

\begin{figure}[!htbp]
\centering
\includegraphics[width=0.65\textwidth]{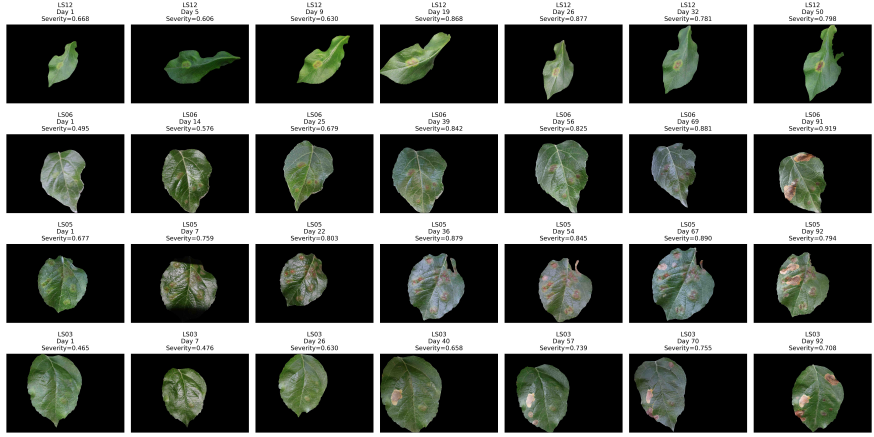}
\vspace{-3mm}
\caption{Representative longitudinal apple scab sequence from leaf LS12, LS06, LS05 and LS03 illustrating disease progression across multiple observation days.}
\label{fig:longitudinal_sequence_example}
\end{figure}

The construction process began by reviewing the complete repository to identify leaves that had been photographed on multiple occasions throughout the growing season. Images corresponding to the same physical leaf were grouped using visual appearance, branch position, acquisition date, and field observations recorded during data collection. Each verified leaf was subsequently assigned a unique Leaf Sequence (LS) identifier, enabling all observations associated with that leaf to be linked into a single temporal sequence. After sequence formation, images were arranged chronologically according to their acquisition dates. Only sequences containing sufficient temporal observations with clearly visible disease progression were retained for further processing. Individual images exhibiting severe motion blur, excessive occlusion, poor focus, duplicate observations, or uncertain leaf identity were discarded to preserve dataset quality and temporal consistency. Figure~\ref{fig:longitudinal_sequence_example} presents an example of a representative apple scab progression sequence.

\begin{figure*}[!htbp]
\centering
\includegraphics[width=0.6\textwidth]{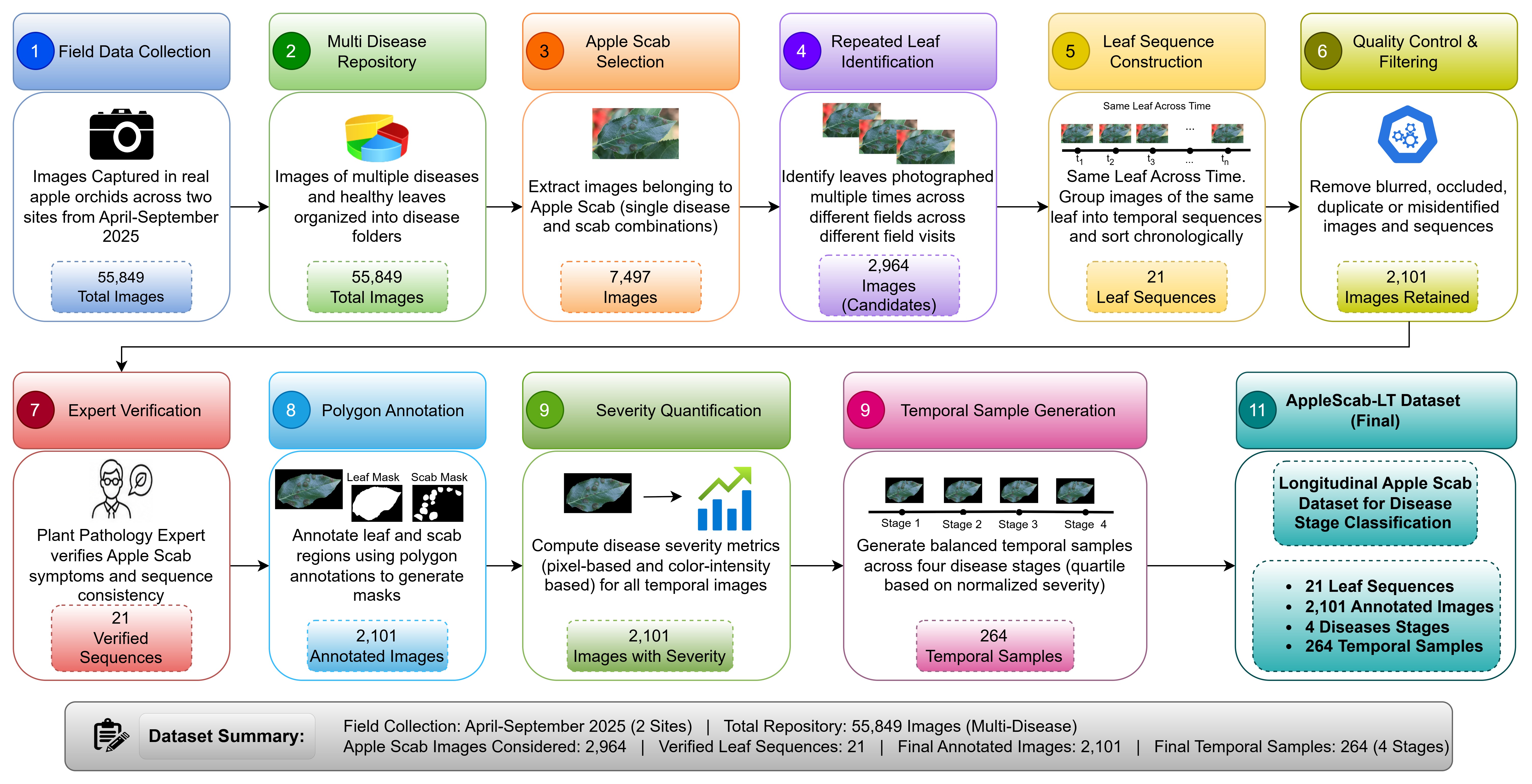}
\vspace{-3mm}
\caption{Construction pipeline of the AppleScab-LT dataset.}
\label{fig:dataset_pipeline}
\end{figure*}

The complete workflow adopted for constructing AppleScab-LT from the original multi-disease repository is presented in Fig.~\ref{fig:dataset_pipeline}. The workflow highlights each stage of the curation pipeline, beginning with field image acquisition and concluding with the generation of a scientifically validated longitudinal dataset. The resulting dataset provides a unique resource for investigating disease progression dynamics and enables the development of progression-aware disease assessment systems that extend beyond conventional static image analysis. The curated sequences subsequently underwent expert disease verification, polygon-based annotation, quality assurance, and disease severity quantification, resulting in a standardized longitudinal suitable for temporal disease analysis.

As shown in Fig.~\ref{fig:dataset_pipeline}, AppleScab-LT was not generated through conventional image collection alone but through a comprehensive curation pipeline incorporating temporal sequence verification, expert validation, standardized annotation, and quantitative severity estimation. These stages collectively ensure that the resulting dataset provides reliable temporal observations suitable for disease progression analysis and temporal plant disease research.

\subsection{Dataset Curation Summary}
\label{sec:Dataset Curation Summary}
Following the identification of repeated apple scab observations, a systematic curation process was performed to transform the original multi-disease repository into the AppleScab-LT dataset. The curation process consisted of disease-specific selection, repeated leaf identification, sequence verification, image quality assessment, expert validation, polygon annotation, and temporal sample generation. Each stage progressively improved the temporal consistency, annotation quality, and scientific reliability of the dataset. The quantitative outcomes of each curation stage are summarized in Table~\ref{tab:curation_summary}. Beginning with the original multi-disease repository, successive curation stages progressively refined the data by identifying apple scab observations, verifying repeated leaf sequences, performing quality control, generating expert annotations, and finally constructing balanced temporal disease-stage samples.

\begin{table}[!htbp]
\centering
\caption{Quantitative summary of the AppleScab-LT dataset curation process.}
\label{tab:curation_summary}
\vspace{-3mm}
\renewcommand{\arraystretch}{0.9}
\small
\begin{tabular}{p{7.5cm}r}
\hline
\textbf{Curation Stage} & \textbf{Count} \\
\hline

Original multi-disease repository & 55,849 images \\

Apple scab images (single + multi-disease) &
7,497 images \\

Candidate repeated observations &
2,964 images \\

Verified leaf sequences &
21 sequences \\

Images retained after quality assessment &
2,101 images \\

Expert verified images &
2,101 images \\

Polygon annotated images &
2,101 images \\

Images with quantified disease severity &
2,101 images \\

Generated temporal samples &
264 samples \\

\hline
\end{tabular}
\end{table}

\begin{figure*}[!htbp]
\centering
\includegraphics[width=0.9\textwidth]{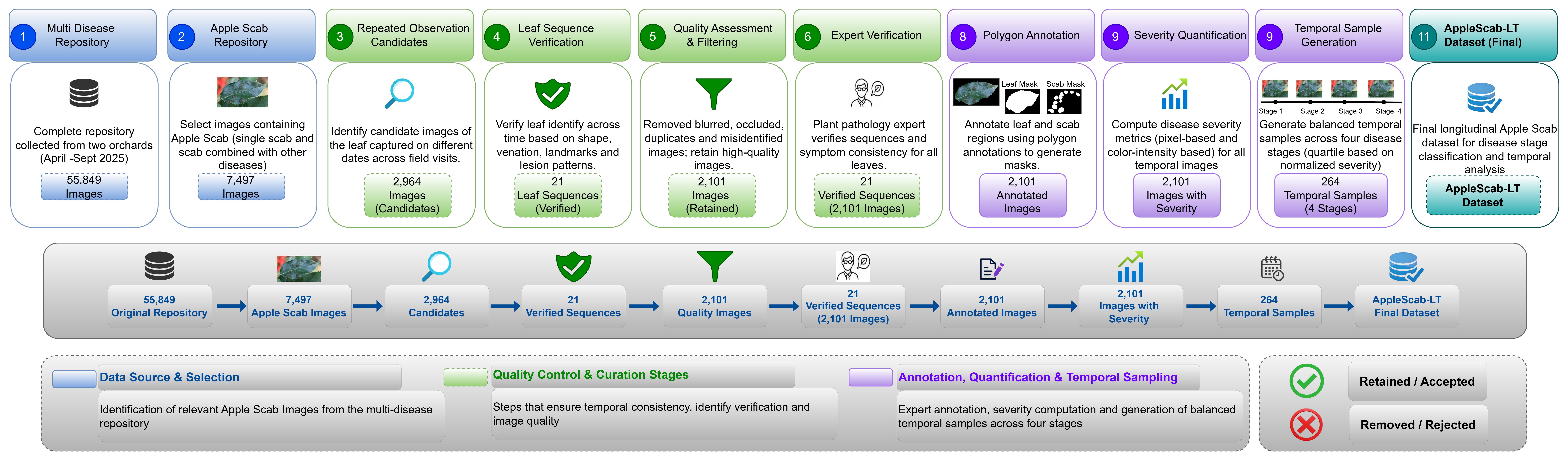}
\vspace{-3mm}
\caption{
AppleScab-LT dataset curation pipeline.}
\label{fig:curation_pipeline}
\end{figure*}

The original repository consisted of 55,849 images acquired under natural orchard conditions. From these, 7,497 images containing apple scab symptoms were identified for further investigation. Subsequent verification of repeated observations resulted in 21 longitudinal leaf sequences comprising 2,101 high-quality images. These curated sequences formed the basis for polygon annotation, disease severity quantification, and the generation of 264 balanced temporal samples used to establish the AppleScab-LT dataset. The complete curation workflow is shown in Fig.~\ref{fig:curation_pipeline}.

As shown in Fig.~\ref{fig:curation_pipeline}, AppleScab-LT was developed through a rigorous multi-stage curation process rather than direct image collection. Each stage contributed to improving the scientific quality of the dataset by ensuring temporal consistency, preserving leaf identity across repeated observations, eliminating low-quality samples, and generating standardized annotations and severity measurements.

\begin{tcolorbox}[colback=blue!5!white,
                  colframe=blue!60!black,
                  title=\textbf{Answer to RQ2},
                  fonttitle=\bfseries,
                  fontupper=\footnotesize]

AppleScab-LT demonstrates that reliable longitudinal disease datasets can be developed through systematic field acquisition, repeated monitoring of individual infected leaves, standardized imaging protocols, and structured dataset organization. The dataset construction framework integrates real orchard acquisition, expert-supported disease verification, sequence-level curation, and standardized metadata generation. Starting from a large-scale field-collected apple disease repository, repeated apple scab observations were identified, validated, and transformed into temporally ordered leaf sequences.

\end{tcolorbox}

\section{RQ3: Annotation, Quality Assurance and Reliability}
\label{sec:annotation}
The scientific value of a high-quality longitudinal dataset depends not only on the quantity of collected images but also on the reliability, consistency, and reproducibility of its annotations. In longitudinal disease monitoring, inaccurate annotations or inconsistent temporal observations may introduce significant errors into disease progression analysis, severity estimation, and temporal modelling. Consequently, ensuring annotation quality becomes considerably more challenging than in conventional static image datasets, where each image is treated independently.

To address the third research question, namely, \emph{``How can annotation quality and temporal reliability be ensured?''}, a comprehensive annotation and quality assurance framework was established for AppleScab-LT. The framework incorporates polygon-based disease delineation, standardized annotation guidelines, expert disease verification, temporal sequence validation, and multiple quality control procedures to ensure both spatial and temporal consistency throughout the dataset. Unlike conventional image-level labels, AppleScab-LT provides detailed pixel-level annotations for both the entire leaf and diseased regions. These annotations enable accurate disease localization, quantitative severity estimation, and reproducible computation of temporal disease progression metrics. Furthermore, repeated observations of individual leaves were independently verified throughout the monitoring period to ensure that every temporal sequence corresponded to the same physical leaf. The following subsections describe the complete annotation methodology together with the procedures adopted to ensure annotation reliability and temporal consistency.

\subsection{Annotation Protocol}
\label{sec:Annotation Protocol}
The primary objective of the annotation process was not only to facilitate disease localization but also to enable accurate estimation of disease severity and longitudinal progression through pixel-level measurements. Image annotation was performed using the open-source LabelMe annotation software, which provides polygon-based semantic annotation capabilities for computer vision datasets. Polygon annotation was selected instead of conventional bounding boxes because apple scab lesions exhibit highly irregular shapes, fragmented symptom regions, and non-uniform lesion boundaries that cannot be accurately represented using rectangular annotations. Precise polygon delineation therefore provides substantially more accurate estimates of lesion area and disease severity. Each image was annotated using two semantic classes:

\begin{itemize}

\item \textbf{Leaf}: A polygon encompassing the complete visible leaf surface. This annotation enables removal of the surrounding background and establishes the total leaf area required for severity computation.

\item \textbf{Scab}: Polygon annotations outlining every visible apple scab lesion present on the leaf surface. Multiple polygons were created whenever several disconnected lesions appeared within the same image.

\end{itemize}

\begin{figure}[!htbp]
\centering
\includegraphics[width=0.5\linewidth]{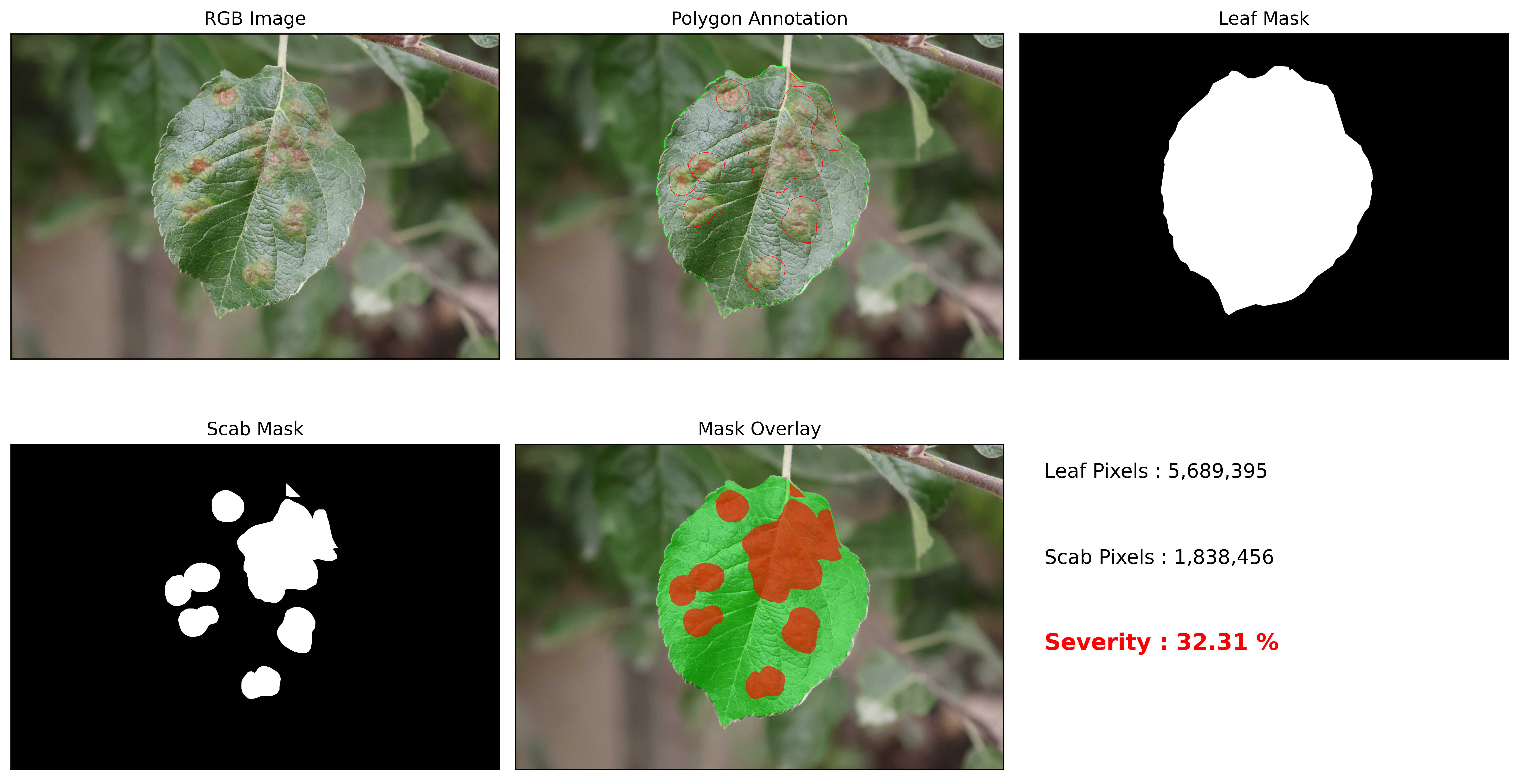}
 \vspace{-3mm}
\caption{Annotation workflow adopted for AppleScab-LT.}
\label{fig:annotation_workflow}
\end{figure}

All annotations were manually generated by carefully tracing lesion boundaries while preserving their natural morphology. During annotation, particular attention was given to lesion edges, partially overlapping symptoms, small isolated lesions, and irregular disease patterns to ensure accurate pixel-level representation. Images exhibiting multiple scab lesions therefore contain multiple independent polygon annotations associated with the same disease class. The complete annotation workflow adopted for AppleScab-LT is illustrated in Fig.~\ref{fig:annotation_workflow}.

To maintain consistency throughout the dataset, identical annotation conventions were applied across all temporal observations belonging to a given leaf sequence. Consequently, changes observed in disease severity arise solely from biological disease progression rather than annotation inconsistencies. As shown in Fig.~\ref{fig:annotation_workflow}, the annotation procedure adopted for AppleScab-LT begins with high-resolution field images that were manually annotated using polygon-based semantic segmentation. Separate polygon annotations were created for the entire leaf and individual scab lesions, followed by automatic generation of binary masks used for pixel-level disease severity estimation and temporal feature extraction. The annotation procedure generates both semantic annotations and corresponding binary masks for each image. These masks provide the foundation for objective disease severity estimation by enabling accurate computation of the proportion of diseased tissue relative to the total visible leaf area. Furthermore, the standardized annotation protocol ensures consistency across repeated observations within each longitudinal leaf sequence.

\subsection{Annotation Guidelines}
\label{sec:Annotation Guidelines}
To ensure annotation consistency throughout the AppleScab-LT dataset, a standardized set of annotation guidelines was established prior to manual labeling. These guidelines were designed to minimize annotation variability and ensure that disease severity measurements accurately reflected biological disease progression rather than inconsistencies introduced during manual annotation. Since AppleScab-LT consists of repeated observations of the same leaf over time, maintaining identical annotation principles across all temporal observations was particularly critical.

All annotations were manually generated using polygon-based semantic segmentation within the LabelMe annotation environment. During annotation, each image was carefully examined individually while simultaneously considering its temporal relationship with other observations belonging to the same leaf sequence. The annotation protocol followed four principal guidelines concerning leaf boundary delineation, disease boundary identification, partial occlusion handling, and treatment of very small lesions.

\subsubsection*{Leaf Boundary Delineation}

The \textit{Leaf} polygon was manually traced along the complete visible boundary of each leaf while excluding all surrounding background pixels. Only visible portions of the leaf surface were annotated. Regions hidden by neighboring leaves, branches, or severe occlusions were not artificially reconstructed. Whenever clearly distinguishable, the natural serrated leaf margins were carefully preserved to ensure accurate estimation of total leaf area required for subsequent disease severity computation.

\subsubsection*{Disease Boundary Delineation}

Apple scab lesions exhibit highly irregular morphology, fragmented symptom regions, and variable lesion sizes. Consequently, every visually distinguishable lesion was independently outlined using one or more polygons depending upon its spatial continuity. Multiple disconnected lesions present on the same leaf were represented using separate polygons associated with the \textit{Scab} class, whereas merged lesions were enclosed within a single polygon encompassing the complete diseased region. Polygon vertices were positioned sufficiently close to lesion boundaries to preserve their natural shape while avoiding unnecessary annotation complexity.

\subsubsection*{Occlusion Handling}

Natural orchard images frequently contain partial occlusions caused by overlapping leaves, branches, shadows, and variations in viewing angle. Only disease symptoms that were directly visible within the captured image were annotated. Lesion regions completely obscured by other objects were intentionally excluded to avoid introducing subjective interpretations regarding invisible disease extent. Images exhibiting severe occlusion that prevented reliable disease assessment were subsequently removed during the quality assurance stage.

\subsubsection*{Tiny Lesion Annotation}

Early-stage apple scab frequently appears as numerous small isolated lesions distributed across the leaf surface. All lesions with clearly identifiable boundaries were manually annotated irrespective of their size. Conversely, extremely small ambiguous spots that could not be confidently distinguished from sensor noise, natural pigmentation, or compression artifacts were excluded from annotation. This strategy reduces false-positive annotations while preserving biologically meaningful disease information essential for temporal progression analysis.

\begin{figure*}[!htbp]
\centering
\includegraphics[width=0.4\textwidth]{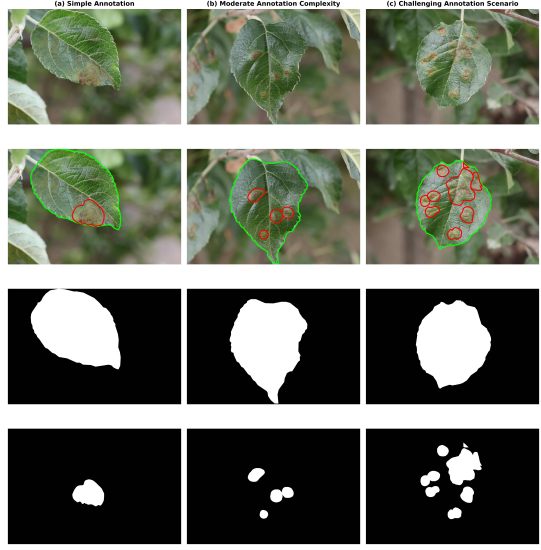}
\caption{
Representative examples illustrating the standardized annotation protocol adopted for AppleScab-LT under varying levels of annotation complexity. Each example presents the original RGB image, polygon annotations, generated leaf mask, and disease mask. (a) \textbf{Simple Annotation},(b) \textbf{Moderate Annotation Complexity}, (c) \textbf{Challenging Annotation Scenario} }
\label{fig:annotation_examples}
\end{figure*}

Collectively, these annotation guidelines establish a standardized protocol for polygon generation throughout the AppleScab-LT dataset. By consistently applying identical annotation principles across all temporal observations, the resulting annotations remain spatially accurate and temporally consistent, thereby providing a reliable foundation for disease localization, severity estimation, temporal progression analysis, and future temporal studies. To illustrate the practical implementation of the proposed annotation guidelines, representative examples corresponding to different levels of annotation complexity are presented in Fig.~\ref{fig:annotation_examples}. These examples demonstrate the application of the standardized annotation protocol under progressively more challenging real-field conditions.

\subsection{Expert Disease Verification and Quality Assurance}
\label{sec:}
To ensure the biological correctness and annotation reliability of AppleScab-LT, every annotated image underwent a comprehensive expert verification procedure following manual polygon annotation. Unlike conventional plant disease datasets that rely solely on a single annotation pass, the proposed longitudinal dataset incorporated a multi-stage verification framework involving independent review by two domain experts before final dataset acceptance.

\begin{figure*}[!t]
\centering
\includegraphics[width=0.59\textwidth]{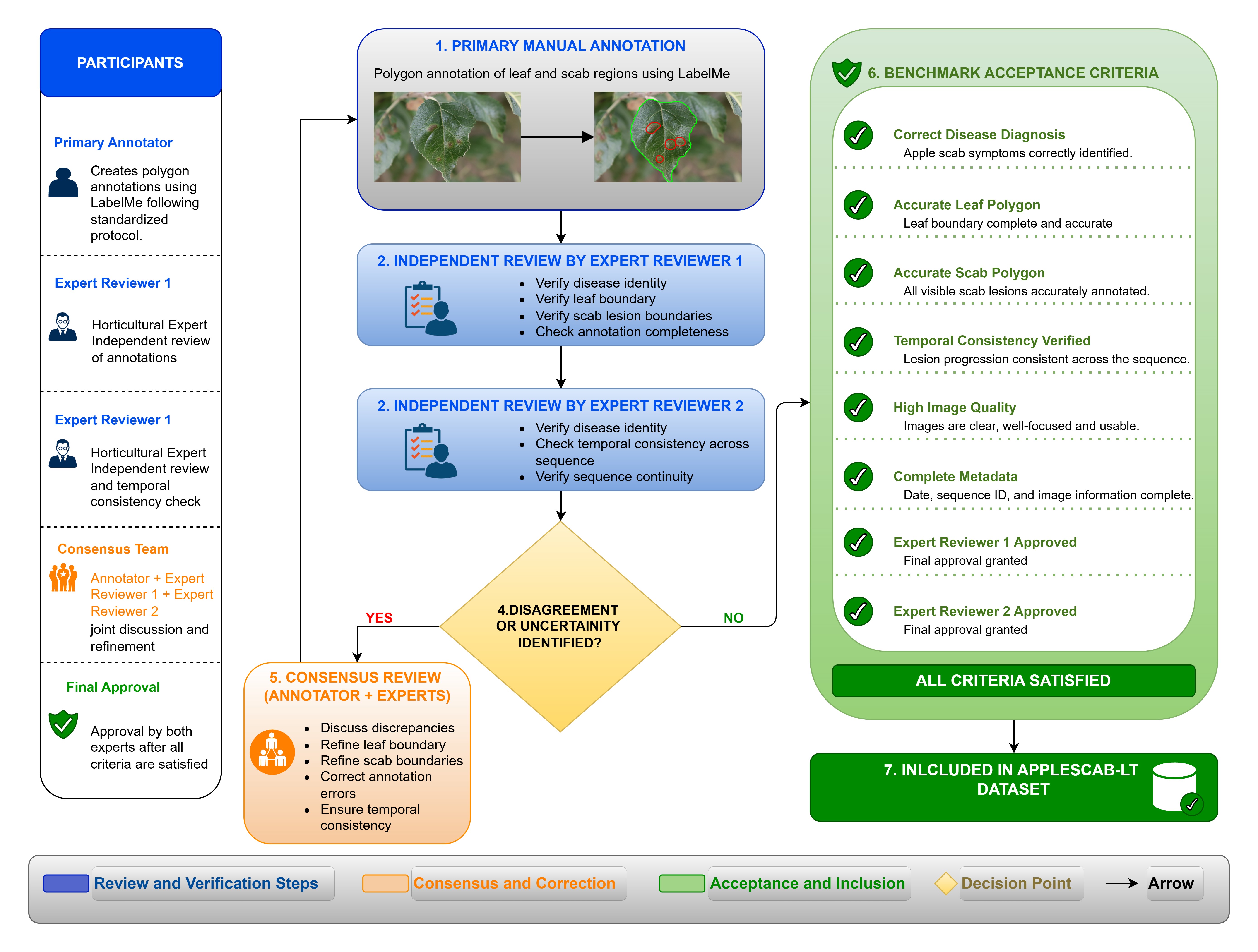}
\vspace{-3mm}
\caption{
Expert disease verification and quality assurance workflow adopted during the construction of the dataset.}
\label{fig:expert_validation}
\end{figure*}

Initially, all polygon annotations were manually generated by the primary annotator according to the standardized annotation guidelines described in Section~\ref{sec:Annotation Guidelines}. Subsequently, two horticultural experts independently reviewed every annotated image. Both experts possess extensive experience in apple disease diagnosis, orchard management, and horticultural pathology under temperate Himalayan growing conditions. The objective of the expert review was to verify disease identity, evaluate annotation accuracy, confirm temporal consistency across repeated observations, and ensure that every annotation accurately represented naturally occurring apple scab symptoms. Whenever disagreements or uncertainties were identified, the annotations were discussed jointly by the primary annotator and both experts until consensus was achieved. The finalized annotations were subsequently incorporated into the final dataset only after expert approval. The complete verification workflow is illustrated in Fig.~\ref{fig:expert_validation}.

\subsubsection*{Primary Annotation}

All images were initially annotated by the primary annotator using polygon-based semantic segmentation. Separate polygons were generated for the complete leaf region and every visible apple scab lesion according to the standardized annotation protocol. During this stage, particular attention was given to accurately delineating irregular lesion boundaries while maintaining consistency across temporal observations belonging to the same leaf sequence.

\subsubsection*{Independent Expert Review}

The two horticultural experts independently evaluated every annotated image without modifying the original annotations during the initial review stage. The evaluation focused on four principal aspects:

\begin{enumerate}
\setlength{\itemsep}{0pt} 
\setlength{\parskip}{0pt}  
\item Correct identification of apple scab symptoms.
\item Accuracy of leaf boundary delineation.
\item Accuracy of scab lesion boundaries.
\item Temporal consistency of disease progression across repeated observations.
\end{enumerate}

Particular attention was given to early-stage lesions, lesion coalescence, irregular disease morphology, and differentiation of apple scab from visually similar disorders including necrosis, and physiological stress symptoms.

\subsubsection*{Iterative Annotation Correction}

Whenever annotation inconsistencies or disease identification errors were detected during expert review, the corresponding images were returned to the primary annotator for correction. Typical revisions included refinement of lesion boundaries, completion of missing disease regions, correction of incomplete leaf polygons, and clarification of ambiguous symptoms. Following correction, the revised annotations were again reviewed by the experts to confirm that all identified issues had been satisfactorily resolved. This iterative review–correction cycle continued until both experts approved the final annotation.

\subsubsection*{Final Dataset Approval}

After successful completion of expert verification and annotation correction, every longitudinal sequence underwent a final quality assessment prior to inclusion within AppleScab-LT dataset. This assessment verified image quality, annotation completeness, sequence continuity, metadata integrity, and temporal consistency.

Only sequences satisfying all predefined quality requirements were retained within the final dataset. Consequently, every image included in AppleScab-LT represents an expert-verified observation supported by standardized annotation guidelines and rigorous quality assurance procedures. As depicted in Fig.~\ref{fig:expert_validation}, the expert verification framework extends beyond conventional annotation review by incorporating predefined acceptance criteria. Every annotated image was required to satisfy all quality requirements before being included in AppleScab-LT. These criteria encompass biological correctness, polygon accuracy, temporal consistency, image quality, metadata integrity, and independent approval by two horticultural experts. The expert verification workflow defines the responsibilities assigned to the primary annotator and the two independent horticultural experts throughout the quality assurance process. To improve transparency and reproducibility, the verification stages together with the dataset acceptance criteria are summarized in Table~\ref{tab:dataset_acceptance_protocol}, which demonstrates that inclusion within AppleScab-LT depended upon satisfying a predefined set of dataset acceptance criteria rather than solely completing manual annotation. This structured verification protocol substantially improves the biological reliability, annotation consistency, and temporal integrity of the proposed longitudinal dataset.

\begin{table*}[!htbp]
\centering
\caption{Verification and acceptance protocol adopted during the construction of the AppleScab-LT dataset.}
\label{tab:dataset_acceptance_protocol}
\vspace{-3mm}
\renewcommand{\arraystretch}{0.9}
\small
\resizebox{\textwidth}{!}{
\begin{tabular}{p{2.0cm}p{3.0cm}p{6.5cm}p{3.2cm}p{4.0cm}}
\hline

\textbf{Verification Stage} &
\textbf{Responsible} &
\textbf{Primary Objective} &
\textbf{Output} &
\textbf{Acceptance Criterion} \\

\hline

Primary Annotation &
Primary Annotator &
Generate polygon annotations for the complete leaf and all visible scab lesions using the standardized annotation protocol. &
Initial annotation set &
Complete leaf and scab polygons generated. \\

\hline

Independent Review I &
Horticultural Expert 1 &
Verify disease identity, lesion morphology, biological correctness, and polygon accuracy. &
Expert review comments &
Disease diagnosis and annotation biologically correct. \\

\hline

Independent Review II &
Horticultural Expert 2 &
Independently verify disease diagnosis, temporal consistency, sequence continuity, and annotation completeness. &
Independent validation report &
Temporal consistency confirmed across repeated observations. \\

\hline

Consensus Review &
Primary Annotator + Both Experts &
Resolve annotation discrepancies through discussion and refinement whenever required. &
Consensus-approved annotation &
All identified inconsistencies resolved. \\

\hline

Final Acceptance &
Both Experts &
Perform final quality assessment before dataset inclusion. &
Final approved annotation &
\begin{tabular}[c]{@{}l@{}}
$\checkmark$ Correct disease diagnosis\\
$\checkmark$ Accurate leaf polygon\\
$\checkmark$ Accurate scab polygon\\
$\checkmark$ Temporal consistency verified\\
$\checkmark$ High image quality\\
$\checkmark$ Complete metadata\\
$\checkmark$ Approved by both experts
\end{tabular}
\\

\hline

\end{tabular}
}
\end{table*}

\subsection{Technical Quality Assurance and Dataset Integrity}
\label{sec:Technical Quality Assurance and Dataset Integrity}
Beyond expert verification of disease annotations, the AppleScab-LT dataset underwent a comprehensive technical quality assurance process to ensure structural integrity, annotation completeness, temporal consistency, and metadata correctness. Since the dataset comprises longitudinal image sequences collected over multiple months, inconsistencies in annotation files, file organization, sequence identifiers, or derived masks could compromise subsequent disease severity estimation and temporal analysis. Consequently, an automated validation framework was developed to systematically examine every dataset component prior to final release. Unlike manual inspection alone, the proposed framework combines validation scripts with targeted visual verification to identify structural inconsistencies that commonly arise during large-scale annotation and dataset curation. The automated validation process substantially reduced manual effort while ensuring reproducibility and consistency throughout the dataset.

\begin{figure*}[!htpb]
\centering
\includegraphics[width=0.43\textwidth]{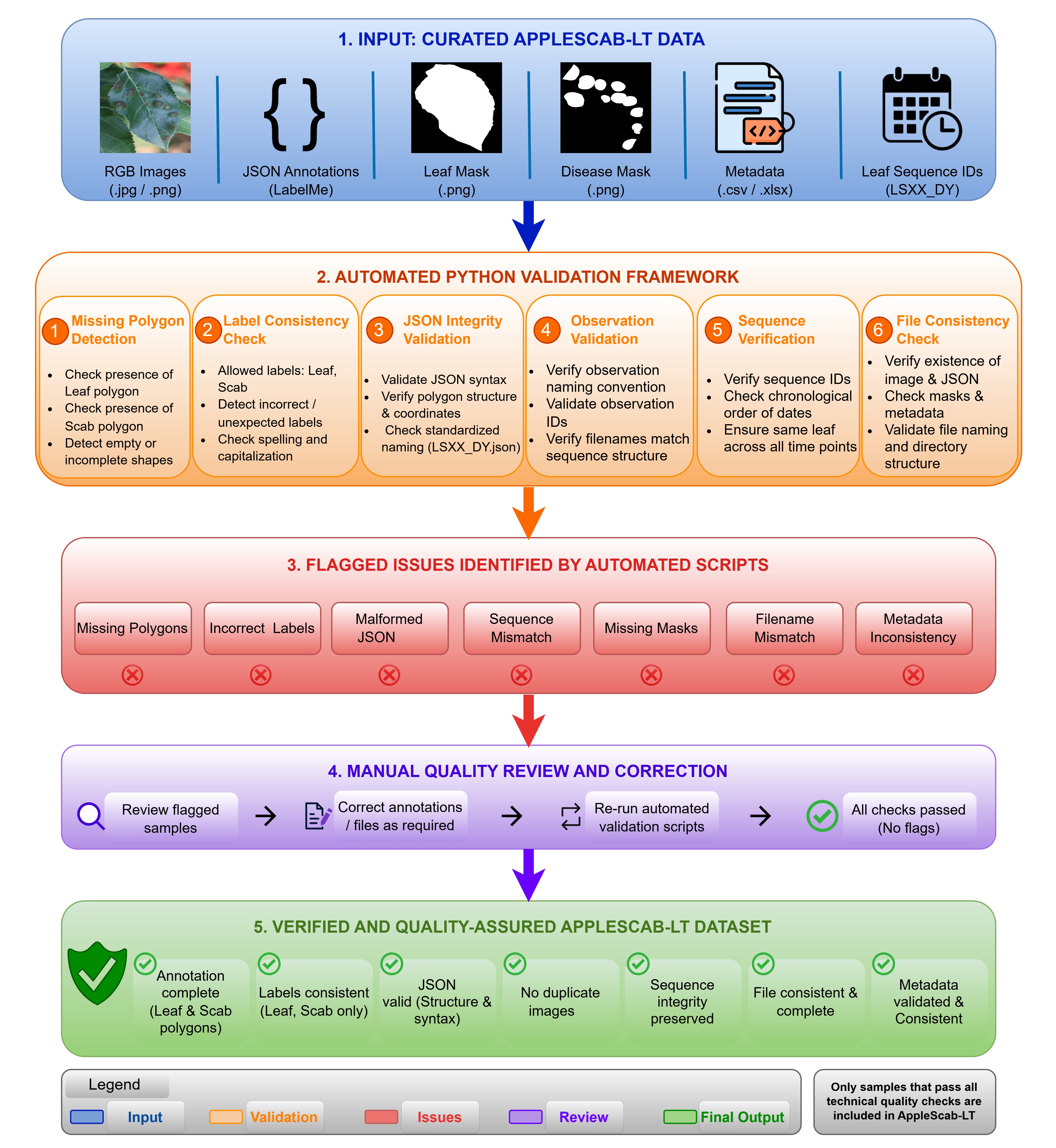}
\vspace{-3mm}
\caption{
Technical quality assurance framework adopted during the construction of AppleScab-LT}
\label{fig:quality_assurance}
\end{figure*}

The validation framework was designed to examine six principal aspects of the dataset, namely polygon completeness, annotation label consistency, JSON integrity, naming verification, temporal sequence verification, and file correspondence. Images or annotations failing any validation stage were automatically flagged for manual inspection, corrected whenever possible, and subsequently re-validated before inclusion within the final dataset.
Figure~\ref{fig:quality_assurance} summarizes the complete technical quality assurance framework adopted during the construction of AppleScab-LT.

\subsubsection*{Missing Polygon Verification}

Validation scripts were used to examine every annotation file for valid polygon annotations corresponding to both semantic classes, namely \textit{Leaf} and \textit{Scab}. The validation process identified annotation files containing missing polygons, incomplete annotations, empty shape definitions, or malformed polygon structures. Images failing these checks were manually reviewed, corrected, and revalidated prior to further processing.

\subsubsection*{Annotation Label Consistency}

To maintain semantic consistency throughout the dataset, validation scripts verified that all annotations exclusively contained the predefined class labels \textit{Leaf} and \textit{Scab}. The scripts additionally detected typographical errors, inconsistent capitalization, duplicated labels, and unexpected annotation classes introduced during manual labeling. Identified inconsistencies were corrected before generating the final annotation files.

\subsubsection*{JSON Integrity Verification}

Since polygon annotations were stored in the LabelMe JSON format, every annotation file underwent automated structural validation. The verification process examined JSON syntax, polygon point integrity, coordinate validity, image dimensions, annotation metadata, and correspondence between each JSON file and its associated RGB image. In addition to structural validation, the framework verified that every annotation file followed the standardized AppleScab-LT naming convention based on leaf sequence identity and observation order (e.g., \texttt{LS01\_D1.json}, \texttt{LS01\_D2.json}, ..., \texttt{LS20\_D7.json}). This ensured that each JSON annotation could be uniquely associated with its corresponding longitudinal observation and prevented inconsistencies arising from incorrect sequence identifiers, duplicated filenames, or mismatched observation indices.

\subsubsection*{Observation Naming Verification}

Since AppleScab-LT preserves multiple observations of the same physical leaf throughout the monitoring period, a standardized naming convention was adopted to uniquely identify every observation. Automated validation scripts verified that all RGB images, JSON annotation files, leaf masks, and disease masks followed the predefined sequence-based naming convention (e.g., \texttt{LS01\_D1 (1)}, \texttt{LS01\_D1 (2)}, \texttt{LS01\_D2 (1)}), thereby ensuring consistent correspondence among all dataset components.

\subsubsection*{Temporal Sequence Verification}

Since AppleScab-LT is fundamentally a longitudinal dataset, preserving the temporal continuity of each leaf sequence represents a critical quality requirement. The validation process confirmed the uniqueness of sequence identifiers, chronological ordering of monitoring days, and completeness of observations within each leaf sequence. Any inconsistencies in filenames, sequence identifiers, or observation indices were automatically flagged and manually corrected prior to including in the dataset.

\subsubsection*{File Consistency Verification}

The final validation stage verified the consistency and completeness of every dataset component associated with each observation. Filename conventions, directory organization, and associated annotation products were simultaneously verified to eliminate missing files, inconsistent naming, and incomplete dataset records.

The workflow demonstrates how automated validation scripts and manual inspection were jointly employed to ensure dataset integrity throughout the quality assurance process. The proposed quality assurance framework combines the efficiency of automated validation with the reliability of expert human supervision to establish the technical integrity of AppleScab-LT. The proposed framework minimizes structural errors and provides a robust foundation for the reliability assessment presented in the following section.

\subsection{Reliability Validation Framework}
\label{sec:Reliability Validation Framework}
The scientific value of a longitudinal dataset depends not only on the accuracy of its annotations but also on the reliability of its temporal observations, structural consistency, and biological correctness. Unlike conventional static image datasets, longitudinal datasets introduce additional sources of variability arising from repeated observations, sequence continuity, annotation consistency, and metadata organization. Consequently, evaluating dataset reliability requires a comprehensive validation strategy extending beyond conventional annotation quality assessment.

To address this challenge, a multidimensional reliability validation framework was established for AppleScab-LT. Rather than representing reliability using a single quantitative metric, the proposed framework evaluates the dataset across four complementary dimensions: (i) spatial annotation reliability, (ii) expert verification reliability, (iii) temporal sequence reliability, and (iv) dataset integrity reliability. Collectively, these dimensions provide a comprehensive assessment of the dataset's suitability for longitudinal disease progression analysis. The overall reliability of AppleScab-LT is defined as

\begin{equation}
R_{\mathrm{AppleScab\text{-}LT}}
=
\left\{
R_S,\;
R_E,\;
R_T,\;
R_D
\right\},
\label{eq:reliability}
\end{equation}
\vspace{-2mm}
where

\begin{itemize}
\setlength{\itemsep}{0pt} 
\setlength{\parskip}{0pt}  

\item $R_S$ denotes the spatial annotation reliability;

\item $R_E$ denotes the expert verification reliability;

\item $R_T$ denotes the temporal sequence reliability;

\item $R_D$ denotes the dataset integrity reliability.

\end{itemize}

\begin{figure*}[!htpb]
\centering
\includegraphics[width=0.5\textwidth]{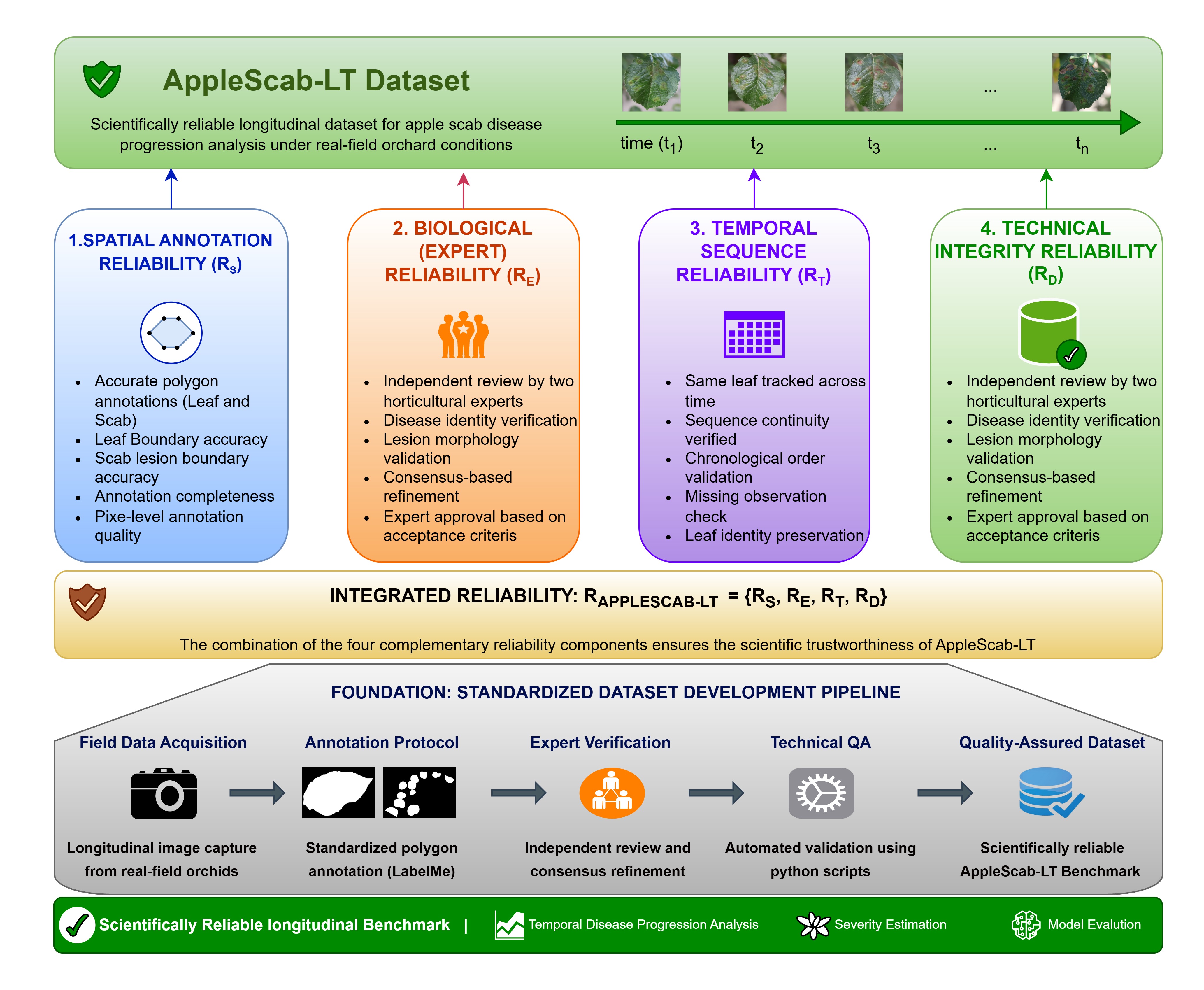}
\vspace{-3mm}
\caption{Proposed reliability validation framework for AppleScab-LT.}
\label{fig:reliability_framework}
\end{figure*}

Rather than representing a single numerical score, Eq.~(\ref{eq:reliability}) defines dataset reliability as a multi-dimensional validation framework in which every component contributes to the overall scientific trustworthiness of the dataset. The reliability framework integrates the annotation protocol, expert review, automated technical validation, and temporal verification procedures presented in the previous sections into a unified validation methodology. Figure~\ref{fig:reliability_framework} presents the proposed reliability validation framework adopted for AppleScab-LT, while Tables~\ref{tab:reliability_dimensions} and \ref{tab:reliability_summary} summarize the individual reliability dimensions and their corresponding validation outcomes.

\vspace{-3mm}
\subsubsection*{Spatial Annotation Reliability}

Spatial annotation reliability refers to the geometric accuracy and semantic consistency of polygon annotations describing the leaf boundary and visible apple scab lesions. The standardized polygon-based annotation protocol described in Section~\ref{sec:Annotation Guidelines} ensured that every annotation accurately represented the visible extent of disease symptoms while preserving irregular lesion morphology.

\subsubsection*{Expert Verification Reliability}

Expert verification reliability was established through independent review performed by two horticultural experts. The experts independently evaluated disease identity, lesion morphology, annotation correctness, and temporal consistency across repeated observations. Whenever discrepancies were identified, annotations underwent consensus-based refinement until agreement was achieved. This iterative review process substantially reduced annotation ambiguity while ensuring biological validity of the dataset.

\subsubsection*{Temporal Sequence Reliability}

Since AppleScab-LT represents repeated observations of individual leaves over time, temporal reliability constitutes one of the defining characteristics of the proposed dataset. Sequence verification confirmed that every observation within a temporal sequence corresponded to the same physical leaf throughout the monitoring period. Sequence identifiers, chronological ordering, acquisition dates, and metadata were independently verified to preserve temporal continuity and ensure that disease progression trajectories accurately reflected biological development rather than annotation inconsistencies or sequence misalignment.

\subsubsection*{Dataset Integrity Reliability}

Dataset integrity reliability evaluates the structural correctness and completeness of every component comprising AppleScab-LT. Automated validation scripts verified annotation completeness, JSON structure, file correspondence, mask availability, metadata integrity, observation verification, filename consistency, and sequence organization. Images failing any validation stage were manually reviewed, corrected where appropriate, and revalidated prior to dataset inclusion. Consequently, the final dataset represents a technically consistent and reproducible resource suitable for future temporal disease analysis.

Collectively, the proposed reliability validation framework establishes AppleScab-LT as a scientifically reliable longitudinal dataset and future progression-aware computer vision research. As presented in Fig.~\ref{fig:reliability_framework}, the scientific reliability of AppleScab-LT is established through the integration of four complementary validation dimensions. Spatial reliability ensures accurate lesion delineation, biological reliability guarantees expert-confirmed disease identification, temporal reliability preserves longitudinal continuity across repeated observations, and technical integrity validates the structural consistency of all dataset components.

\begin{table*}[!htbp]
\centering
\caption{Reliability validation dimensions adopted for AppleScab-LT.}
\label{tab:reliability_dimensions}
\vspace{-3mm}
\renewcommand{\arraystretch}{0.9}
\small
\begin{tabular}{p{4.3cm}p{5.5cm}p{5.8cm}}
\hline
\textbf{Reliability Dimension} &
\textbf{Validation Method} &
\textbf{Purpose} \\
\hline

Spatial Annotation Reliability &
Polygon annotation protocol, expert review, automated polygon validation &
Ensure accurate delineation of leaf and scab regions. \\

Expert Verification Reliability &
Independent review by two horticultural experts and consensus-based refinement &
Ensure biological correctness of disease annotations. \\

Temporal Sequence Reliability &
Sequence verification, chronological ordering, identity preservation &
Ensure continuity of repeated observations of the same leaf. \\

Dataset Integrity Reliability &
Automated Python validation scripts and manual inspection &
Ensure structural consistency, metadata integrity, and dataset completeness. \\

\hline
\end{tabular}
\end{table*}

\vspace{-3mm}
\begin{table*}[!htbp]
\centering
\caption{Summary of reliability validation outcomes for AppleScab-LT.}
\label{tab:reliability_summary}
\vspace{-3mm}
\renewcommand{\arraystretch}{0.9}
\small
\begin{tabular}{p{4cm}p{7.8cm}p{1.5cm}}
\hline
\textbf{Validation Aspect} &
\textbf{Verification Procedure} &
\textbf{Outcome} \\
\hline

Annotation Quality &
Standardized polygon annotation and expert review &
Verified \\

Disease Identity &
Independent review by two horticultural experts &
Verified \\

Temporal Continuity &
Leaf sequence verification and chronological validation &
Verified \\

Dataset Integrity &
Automated Python validation and manual inspection &
Verified \\

Metadata Consistency &
Cross-validation of filenames, sequence IDs, masks, and JSON files &
Verified \\

Dataset Validation &
Compliance with predefined acceptance criteria &
Verified \\

\hline
\end{tabular}
\end{table*}
\vspace{-3mm}

\begin{tcolorbox}[colback=blue!5!white,
                  colframe=blue!60!black,
                  title=\textbf{Answer to RQ3},
                  fonttitle=\bfseries,
                  fontupper=\footnotesize]

AppleScab-LT ensures annotation quality and temporal reliability through a multi-stage validation framework specifically designed for longitudinal disease datasets. The framework combines standardized polygon-based annotation, expert disease verification, consensus-based refinement, automated technical quality assurance, and predefined acceptance criteria to ensure that every observation satisfies spatial, biological, and structural reliability requirements. Beyond conventional image-level validation, the proposed reliability framework explicitly verifies temporal consistency by preserving leaf identity, chronological sequence ordering, and continuity across repeated observations. Automated validation procedures further ensure completeness, consistency, and correctness of the dataset. 

\end{tcolorbox}

\section{RQ4: Dataset Characterization}
\label{sec:dataset_characterization}
Following dataset construction and comprehensive reliability validation, the proposed AppleScab-LT dataset was quantitatively characterized to summarize its structural properties, temporal organization, and disease progression coverage. Dataset characterization serves two principal objectives. First, it provides a comprehensive overview of the dataset contents, including image statistics, annotation resources, and temporal observations. Second, it demonstrates the diversity and completeness of the longitudinal sequences, thereby enabling the scope and applicability of the dataset for disease progression modelling. AppleScab-LT is organized around repeated observations of individual leaves rather than isolated images and is characterized not only by the number of images collected but also by the temporal structure of the monitored sequences. The following subsections summarize the principal characteristics of the dataset and provide quantitative evidence regarding its longitudinal composition.

\subsection{Overall Dataset Statistics}
\label{sec:Overall Dataset Statistics}
Table~\ref{tab:overall_statistics} summarizes the overall characteristics of the AppleScab-LT dataset. The dataset comprises longitudinal observations collected from twenty one independently monitored leaf sequences under natural orchard conditions. Each observation includes the original RGB image together with polygon annotations, binary masks, and metadata required for quantitative disease progression analysis. In addition to image-level annotations, every observation is associated with corresponding LabelMe JSON annotation files, leaf masks, disease masks, and temporal metadata describing sequence identity and acquisition chronology.

\begin{table}[!htbp]
\centering
\caption{Overall characteristics of the AppleScab-LT longitudinal dataset.}
\vspace{-3mm}
\label{tab:overall_statistics}
\renewcommand{\arraystretch}{0.9}
\small
\begin{tabular}{lll}
\hline
\textbf{Category} & \textbf{Attribute} & \textbf{Value} \\
\hline

\multirow{2}{*}{Dataset}
& Total RGB images & 2,101 \\
& Leaf sequences & 21 \\

\hline

Annotations
& JSON files, Leaf Masks, Disease Masks, Polygon Annotations & 2,101 \\

\hline

\multirow{3}{*}{Temporal}
& Temporal observations & 2,101 \\
& Disease stages & 4 \\
& Generated temporal samples & 264 \\

\hline
\end{tabular}
\end{table}

\subsection{Longitudinal Sequence Characteristics}
\label{sec:Longitudinal Sequence Characteristics}
The defining characteristic of AppleScab-LT is its organization into longitudinal leaf sequences rather than independent disease images. Each monitored leaf was repeatedly observed throughout the disease development period, resulting in temporal sequences exhibiting variable durations and observation frequencies depending upon disease progression and field conditions. AppleScab-LT was derived from a larger field-acquired multi-disease apple repository collected under natural orchard conditions. The resulting monitoring duration therefore varies naturally among leaf sequences depending upon the infection onset time, disease progression rate, leaf senescence, environmental conditions, physical occlusion, and continued visibility of the monitored leaf. This acquisition strategy preserves realistic disease development rather than enforcing an artificially uniform monitoring schedule. Table~\ref{tab:leaf_statistics} summarizes the characteristics of each monitored leaf sequence, including the number of temporal observations, monitoring duration, and sequence length.

\begin{table*}[!htbp]
\centering
\caption{Leaf-wise longitudinal characteristics of the AppleScab-LT dataset.}
\label{tab:leaf_statistics}
\vspace{-3mm}
\resizebox{\textwidth}{!}{%
\begin{tabular}{lccccc}
\hline
\textbf{Leaf} &
\textbf{Monitoring Duration (Days)} &
\textbf{Temporal Observations} &
\textbf{Captured Images} &
\textbf{Mean Interval (Days)} &
\textbf{Images/Observation} \\
\hline
LS01 & 30 & 12 & 37 & 2.73 & 3.08 \\
LS02 & 92 & 9 & 92 & 11.50 & 10.22 \\
LS03 & 92 & 28 & 184 & 3.41 & 6.57 \\
LS04 & 38 & 17 & 131 & 2.38 & 7.71 \\
LS05 & 92 & 33 & 417 & 2.88 & 12.64 \\
LS06 & 91 & 30 & 325 & 3.14 & 10.83 \\
LS07 & 54 & 13 & 62 & 4.50 & 4.77 \\
LS08 & 55 & 10 & 74 & 6.11 & 7.40 \\
LS09 & 28 & 10 & 54 & 3.11 & 5.40 \\
LS10 & 24 & 9 & 64 & 3.00 & 7.11 \\
LS11 & 20 & 9 & 62 & 2.50 & 6.89 \\
LS12 & 50 & 21 & 102 & 2.50 & 4.86 \\
LS13 & 40 & 9 & 57 & 5.00 & 6.33 \\
LS14 & 10 & 4 & 43 & 3.33 & 10.75 \\
LS15 & 54 & 5 & 52 & 13.50 & 10.40 \\
LS16 & 66 & 9 & 75 & 8.25 & 8.33 \\
LS17 & 61 & 13 & 86 & 5.08 & 6.62 \\
LS18 & 67 & 7 & 65 & 11.17 & 9.29 \\
LS19 & 12 & 4 & 38 & 4.00 & 9.50 \\
LS20 & 42 & 7 & 51 & 7.00 & 7.29 \\
LS21 & 73 & 5 & 30 & 18.25 & 6.00 \\
\hline
\end{tabular}
}
\end{table*}

Table~\ref{tab:leaf_statistics} shows the diversity of AppleScab-LT in terms of monitoring duration, temporal sampling frequency, and image acquisition density. The monitoring duration varies from 9 to 91 days because leaf sequences entered the longitudinal monitoring process at different stages of the overall field campaign. Consequently, some sequences contain only a few temporal observations, whereas others comprise extended monitoring trajectories spanning more than thirty observations. Rather than representing missing data, this variability reflects realistic orchard monitoring conditions and captures the natural heterogeneity of disease development across individual leaves.

\begin{figure*}[!htbp]
\centering
\includegraphics[width=0.4\textwidth]{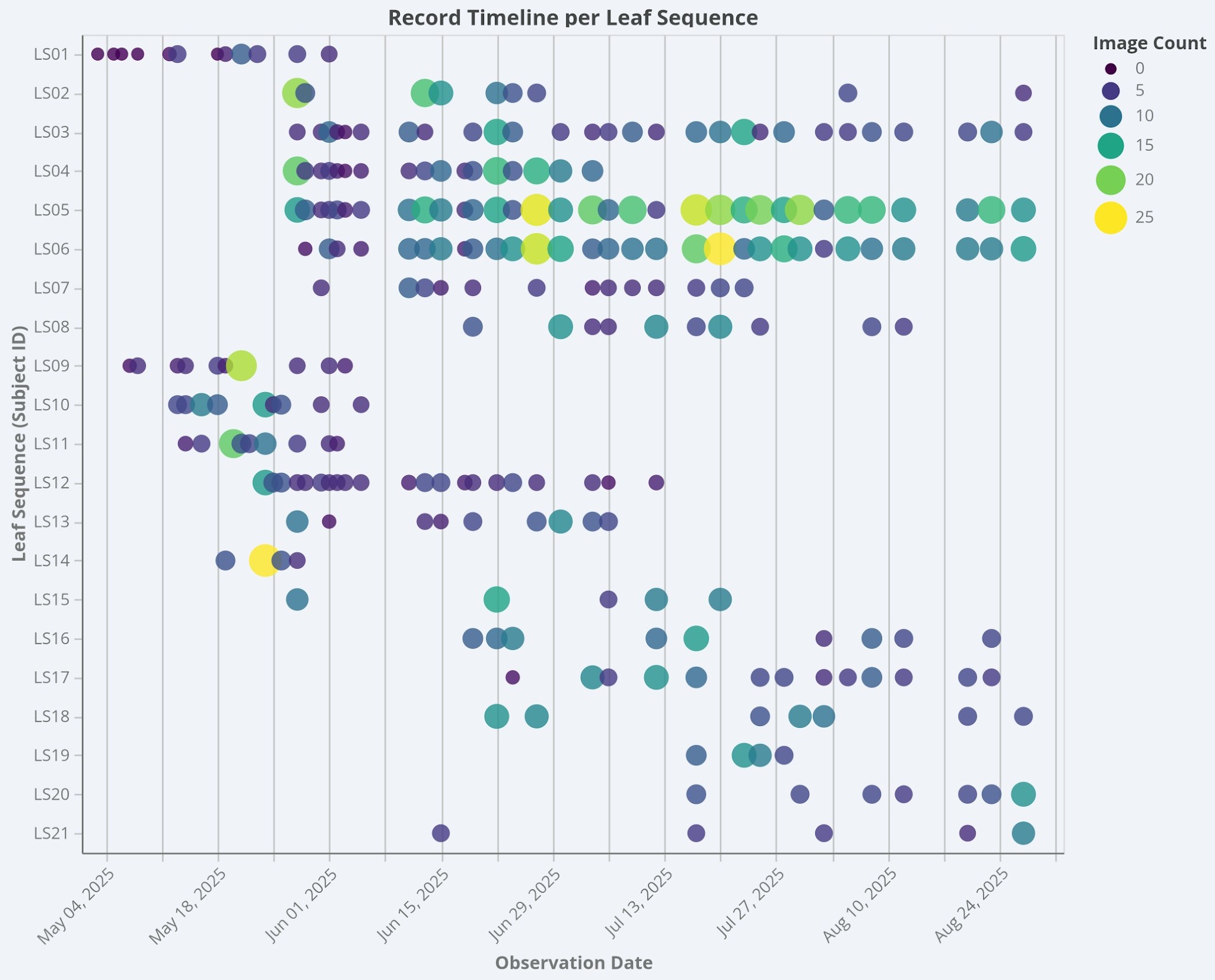}
\vspace{-3mm}
\caption{Distribution of temporal sequence lengths in the proposed AppleScab-LT dataset.}
\label{fig:record_timeline}
\end{figure*}

Figure~\ref{fig:record_timeline} shows the record timeline of temporal sequence across all monitored leaves. The observed variability reflects natural differences in disease progression, monitoring duration, and field acquisition conditions. The dataset contains variable-length longitudinal trajectories ranging from four to thirty-three temporal observations. This variability arises from the practical realities of long-term field monitoring rather than experimental design. Since AppleScab-LT was extracted from a comprehensive multi-disease apple repository, different infected leaves entered the monitoring process at different time points and remained observable for varying durations. Furthermore, environmental factors such as weather conditions, leaf senescence, partial occlusion, camera accessibility, and differences in disease progression rates naturally influenced the number of observations obtained for each sequence. Preserving these unequal sequence lengths provides a realistic representation of orchard monitoring and establishes a challenging task for temporal disease intelligence to learning disease evolution without requiring uniformly sampled sequences.

\subsection{Disease Severity Quantification}
\label{sec:Disease Severity Quantification}
An important characteristic of AppleScab-LT is the availability of quantitative disease severity measurements for every temporal observation. Unlike conventional plant disease datasets that provide only categorical disease labels, the proposed dataset incorporates multiple complementary severity descriptors that characterize disease development throughout the monitoring period.

Disease severity was computed from the polygon-based annotations generated during the segmentation process. Separate leaf and disease masks enabled pixel-level estimation of infected leaf area, color-based measurements provided additional information regarding lesion appearance and symptom evolution, and relative severity provided the progression of severity over the mionitoring period. The resulting severity descriptors constitute an important component of AppleScab-LT and provide continuous quantitative information that complements the visual observations. These measurements can be directly employed for disease progression modelling, severity estimation, temporal forecasting, and progression-aware machine learning.

\subsubsection*{Pixel-based Severity}

Pixel-based disease severity represents the proportion of diseased tissue relative to the total visible leaf area. Following generation of the binary leaf and disease masks, the number of pixels belonging to each class was computed automatically. Disease severity was subsequently estimated as

\begin{equation}
S_{pixel}
=
\frac{N_{disease}}
{N_{leaf}}
\times100,
\label{eq:pixelseverity}
\end{equation}

where $N_{disease}$ denotes the number of pixels contained within the disease mask and $N_{leaf}$ represents the total number of pixels contained within the corresponding leaf mask.

\subsubsection*{Color-based Severity}

Disease progression is accompanied not only by expansion of lesion area but also by progressive changes in lesion colour and intensity. Consequently, AppleScab-LT additionally incorporates a color-based severity descriptor computed from the segmented disease regions. Average colour intensity was extracted from every lesion using the corresponding disease mask, thereby providing complementary information regarding symptom development that cannot be captured solely through lesion area measurements.

\subsubsection*{Normalized Relative Severity}

To facilitate comparison among longitudinal leaf sequences exhibiting different maximum disease severities, a normalised relative severity measure is used. The disease severity of each observation is expressed relative to the maximum pixel-based severity observed within the corresponding leaf sequence. The normalized relative severity is computed as

\begin{equation}
S_{NRS}
=
\frac{S_{pixel,i}}
{\max(S_{pixel})},
\label{eq:nrs}
\end{equation}

where $S_{pixel,i}$ denotes the pixel-based disease severity at the $i^{th}$ temporal observation, and $\max(S_{pixel})$ represents the maximum pixel-based severity recorded for that leaf sequence. 

\begin{figure*}[!htbp]
\centering
\includegraphics[width=0.9\textwidth]{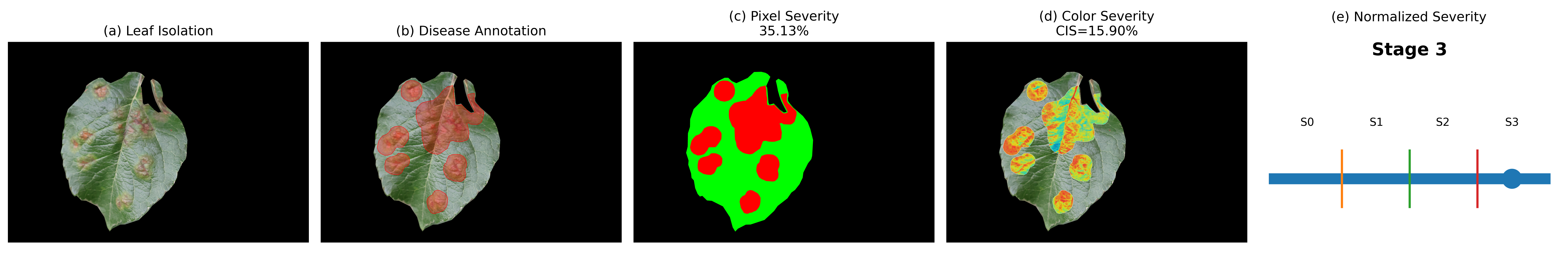}
\vspace{-2mm}
\caption{
Disease severity quantification framework adopted in AppleScab-LT.}
\label{fig:severity_computation}
\end{figure*}

This sequence-wise normalization preserves the relative progression dynamics of each monitored leaf while enabling meaningful comparison among different longitudinal trajectories despite differences in absolute lesion size. Figure~\ref{fig:severity_computation} illustrates the complete disease severity computation framework adopted in AppleScab-LT, beginning with polygon annotations and concluding with quantitative severity descriptors suitable for longitudinal disease analysis.

\subsection{Longitudinal Disease Progression Dynamics}
\label{sec:Longitudinal Disease Progression Dynamics}
\begin{figure*}[!htbp]
\centering
\includegraphics[width=0.5\textwidth]{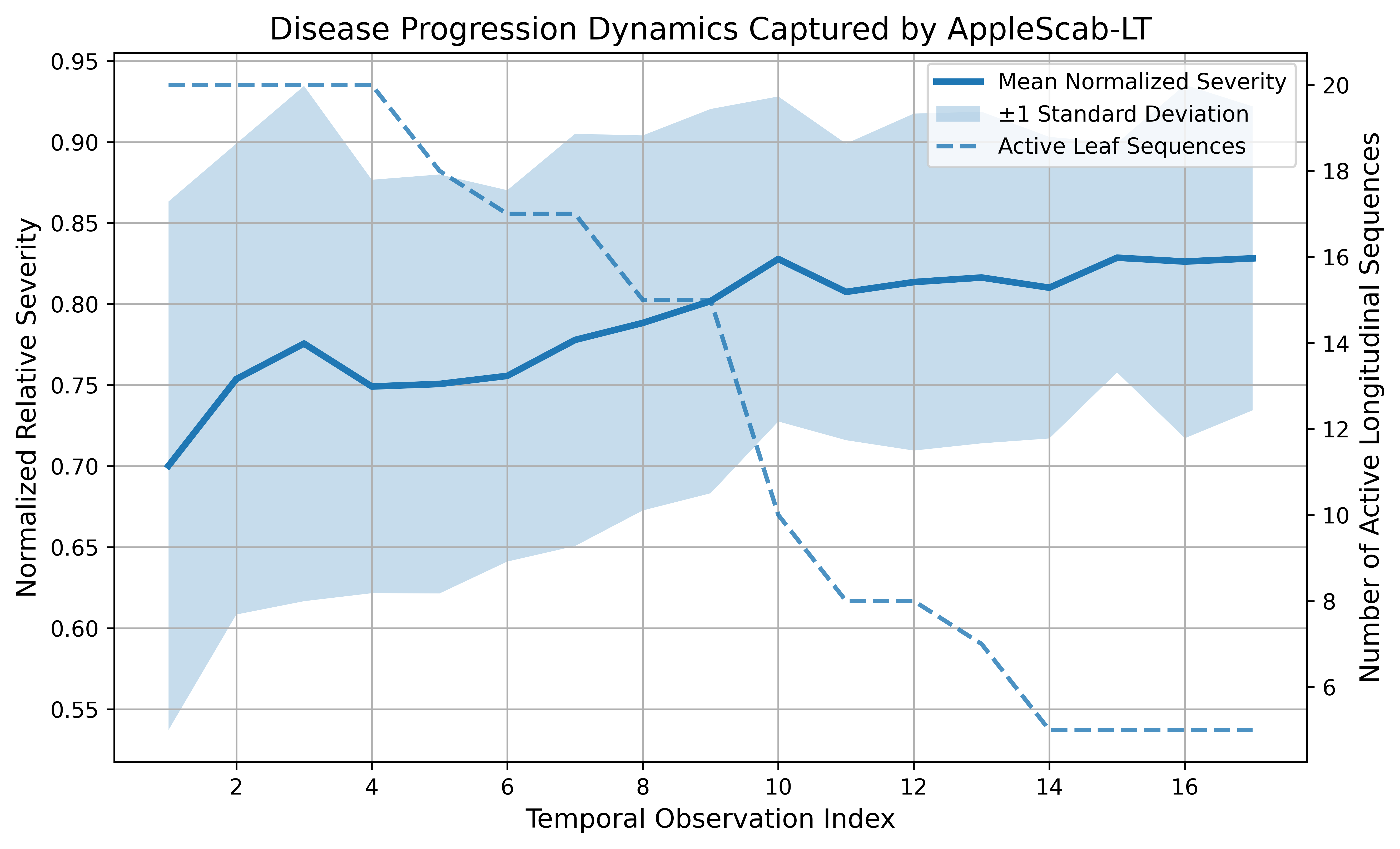}
\vspace{-3mm}
\caption{Longitudinal disease progression dynamics captured by AppleScab-LT.}
\label{fig:progression_curve}
\end{figure*}

The principal contribution of AppleScab-LT lies in its ability to capture the temporal evolution of apple scab under natural orchard conditions. A defining characteristic of AppleScab-LT is its ability to capture the continuous temporal evolution of apple scab under natural orchard conditions. Unlike conventional plant disease datasets that provide isolated snapshots of disease symptoms, AppleScab-LT records repeated observations of the same physical leaf throughout its disease development, thereby preserving the biological progression of infection over time. This longitudinal monitoring enables quantitative investigation of disease onset, lesion expansion, progression rate, and severity evolution using real-field observations.
To characterize the overall disease progression behaviour, the normalized disease severity trajectories of all monitored leaf sequences were aligned with respect to the first observation of each sequence. This temporal normalization allows disease progression to be analysed on a common biological timescale despite differences in acquisition dates and monitoring durations among individual leaves. Figure~\ref{fig:progression_curve} summarizes the longitudinal disease progression dynamics captured by AppleScab-LT. The solid curve represents the mean normalized relative disease severity computed across all active leaf sequences at each temporal observation, while the shaded region denotes one standard deviation, illustrating the natural variability in disease progression under real-field conditions. The dashed curve indicates the number of active longitudinal sequences contributing to the mean severity estimate at each observation. The gradual decline in active sequences reflects differences in monitoring duration among individual leaves due to natural senescence, field accessibility, and completion of monitoring, rather than missing observations. Collectively, these results demonstrate that AppleScab-LT captures both the average progression behaviour and the inherent biological variability associated with apple scab development.

Although individual leaf sequences exhibit different progression rates, disease onset times, and final severity levels, a consistent increasing trend in normalized disease severity is evident throughout the monitoring period. Such variability reflects the influence of natural environmental conditions, host responses, microclimatic variability, and biological differences among individual leaves rather than inconsistencies in data collection. Consequently, the dataset preserves realistic disease heterogeneity while maintaining temporal continuity across successive observations. The availability of longitudinal severity trajectories distinguishes AppleScab-LT from existing static plant disease datasets and substantially extends its potential applications beyond conventional image classification , thereby directly addressing Research Question~4 by providing a longitudinal dataset capable of representing temporal disease characteristics. The dataset provides a reliable foundation for disease progression modelling, severity forecasting, temporal representation learning, progression-aware deep learning, disease stage classification, early disease detection, and future temporal disease intelligence systems.

\subsection{Inter-Leaf Disease Variability}
\label{sec:Inter-Leaf Disease Variability}
\begin{figure*}[!htbp]
\centering
\includegraphics[width=0.6\textwidth]{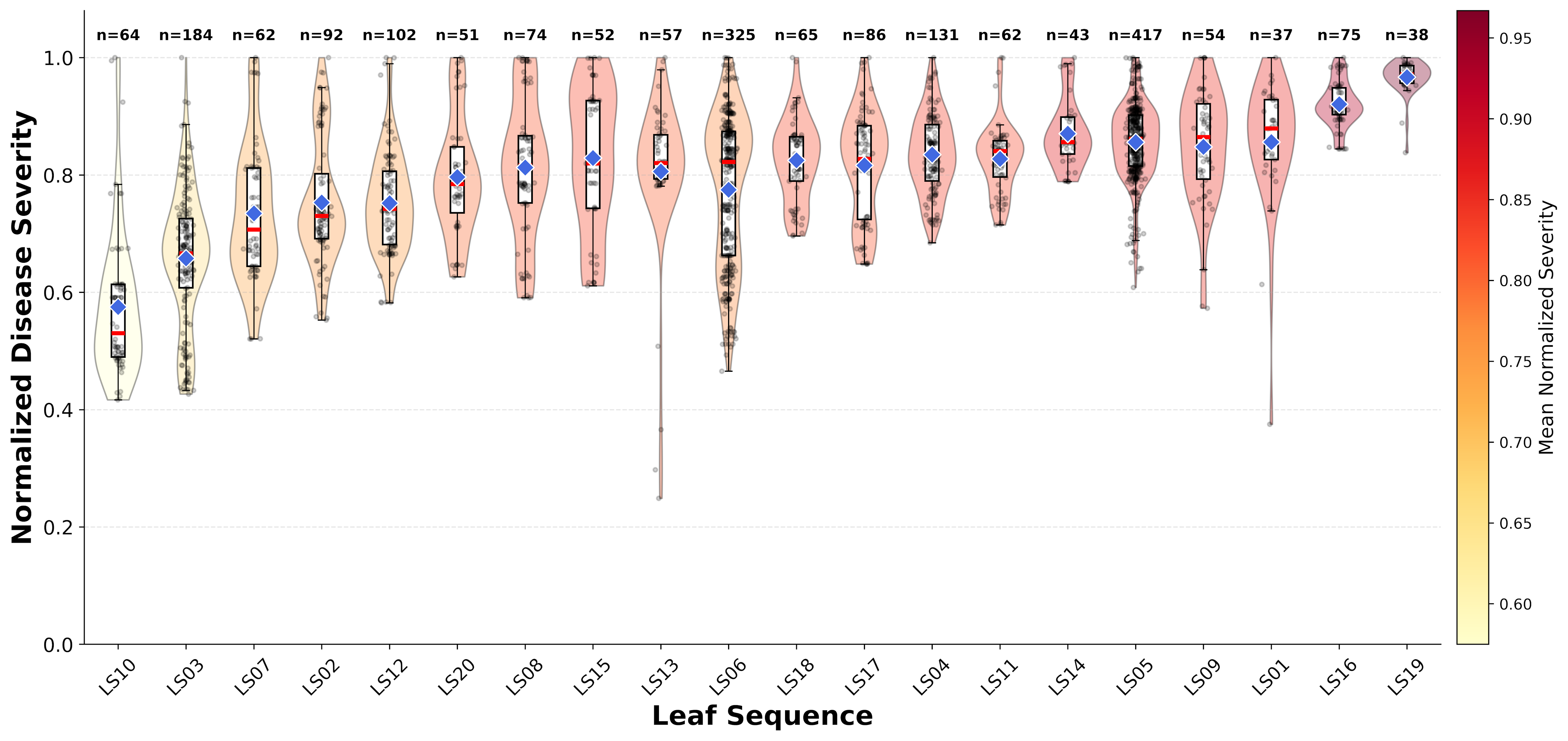}
\vspace{-3mm}
\caption{
Inter-leaf variability in normalized disease severity across the AppleScab-LT dataset.}
\label{fig:interleaf_variability}
\end{figure*}

An important characteristic of AppleScab-LT is its ability to capture the natural variability of disease progression among different leaves exposed to real-field orchard conditions. Disease development in commercial orchards is influenced by numerous biological and environmental factors, including leaf age, canopy position, microclimatic conditions, pathogen inoculum density, and host susceptibility. Consequently, substantial variation in disease severity is expected even among leaves monitored within the same orchard. To investigate this variability, the distribution of normalized disease severity was analysed separately for each monitored leaf sequence. Figure~\ref{fig:interleaf_variability} presents boxplots summarizing the statistical distribution of disease severity across all temporal observations belonging to each leaf sequence. The boxplots illustrate that each violin represents the distribution of normalised disease severity observations collected from an individual leaf sequence throughout the monitoring period. Embedded boxplots summarize the median and interquartile range, while individual points correspond to temporal observations. Blue diamond markers indicate the mean severity of each sequence, and the accompanying colour scale represents the average normalized disease severity. Leaf sequences are arranged in ascending order of median disease severity, illustrating the broad spectrum of disease progression behaviours captured under natural orchard conditions.

The observed variability demonstrates that AppleScab-LT preserves realistic biological heterogeneity rather than enforcing artificially uniform disease progression patterns. Some leaf sequences exhibit relatively slow and gradual disease development, whereas others display rapid lesion expansion accompanied by greater severity variation. Such diversity represents an important characteristic of longitudinal field datasets and enables robust evaluation of disease progression models under realistic conditions. The inter-leaf variability analysis confirms that the dataset captures a broad spectrum of disease progression behaviours while maintaining temporal consistency within individual sequences.

\subsection{Temporal Disease Behaviour Analysis}
\label{sec:Temporal Disease Behaviour Analysis}
\begin{figure*}[!htbp]
\centering
\includegraphics[width=0.55\textwidth]{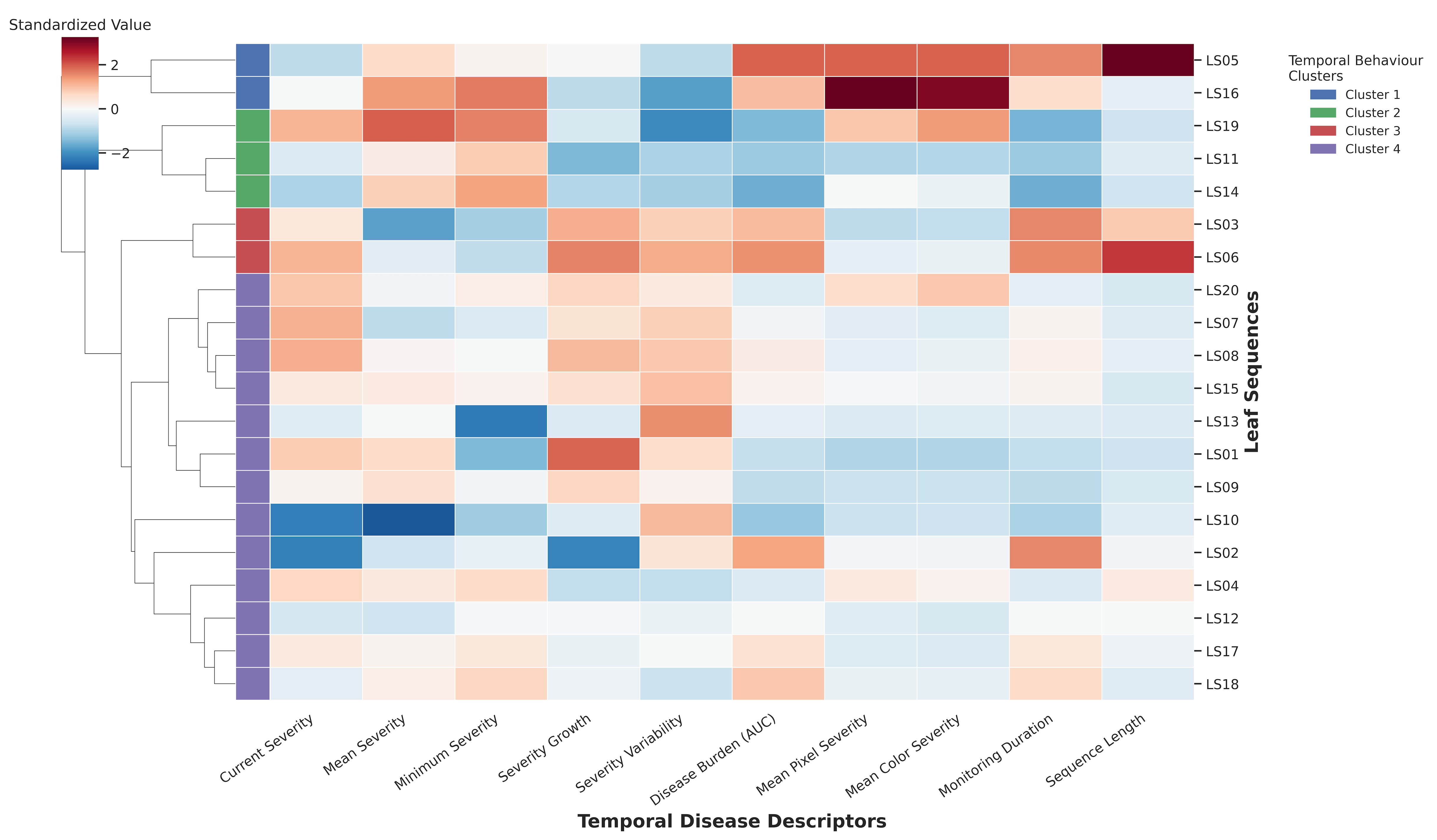}
\caption{
Temporal disease descriptor map of AppleScab-LT generated using ten longitudinal characterization descriptors extracted from each monitored leaf sequence.
}
\label{fig:descriptor_map}
\end{figure*}

A key objective of AppleScab-LT is to capture not only disease severity at individual observation points but also the diverse temporal behaviours exhibited during disease progression under natural orchard conditions. To characterize these longitudinal patterns, ten temporal disease descriptors were extracted from each monitored leaf sequence, encompassing disease status, progression dynamics, visual severity characteristics, and monitoring attributes. Unlike conventional image datasets that provide only static disease labels, AppleScab-LT enables comprehensive characterization of disease evolution throughout the monitoring period. Figure~\ref{fig:descriptor_map} presents a temporal disease descriptor map generated from standardized descriptor values using hierarchical clustering. Each row corresponds to an individual leaf sequence, while each column represents one temporal descriptor. Standardization enables direct comparison among descriptors with different numerical ranges, while hierarchical clustering groups leaf sequences exhibiting similar temporal characteristics.

The descriptor map reveals considerable heterogeneity in disease progression across the monitored leaf sequences. Several groups of leaves exhibit consistently higher disease burden, larger visual severity measurements, and longer monitoring periods, whereas other groups are characterized by lower disease severity, reduced progression, or shorter observation histories. Importantly, these clusters are generated in an unsupervised manner and therefore represent naturally occurring similarities in temporal disease behaviour rather than predefined disease categories. This observation demonstrates that AppleScab-LT captures multiple progression phenotypes that arise under real-field orchard conditions.

The diversity observed across the temporal descriptors highlights one of the principal strengths of the proposed dataset. Disease progression is influenced by numerous biological and environmental factors, including host susceptibility, pathogen development, microclimatic variation, and lesion expansion dynamics. Consequently, disease evolution does not follow a single uniform trajectory. Instead, AppleScab-LT preserves a broad spectrum of temporal progression behaviours, providing realistic variability that is essential for the development of robust progression-aware artificial intelligence systems.

\subsection{Disease Stage Construction}
\label{sec:Disease Stage Construction}
Although disease severity evolves continuously throughout the infection process, many downstream computer vision tasks, including disease stage classification, require discrete stage labels. To facilitate disease stage construction while preserving the temporal characteristics of disease progression, AppleScab-LT employs a data-driven quartile-based disease stage construction strategy. For each monitored leaf sequence, normalized relative disease severity was computed from the segmented disease and leaf regions. The complete distribution of severity values obtained from all longitudinal observations was subsequently partitioned using the first quartile (Q1), median (Q2), and third quartile (Q3). These statistically derived thresholds divide the continuous severity distribution into four balanced disease stages representing progressively increasing disease development. 

Unlike manually selected thresholds, quartile-based partitioning automatically adapts to the empirical severity distribution of the dataset while producing approximately balanced stage populations. The use of quartiles ensures that each disease stage contains a comparable number of observations, thereby reducing class imbalance and enabling statistically fair evaluation of disease stage classification algorithms. Furthermore, because the thresholds are derived directly from the observed severity distribution, the resulting stage definitions remain objective, reproducible, and independent of expert-defined severity scales.

Let $S=\{s_1,s_2,\ldots,s_n\}$ denote the normalized relative severity values obtained from all longitudinal observations. The quartile thresholds are computed as

\begin{equation}
Q_1=P_{25}(S), \qquad
Q_2=P_{50}(S), \qquad
Q_3=P_{75}(S),
\end{equation}

where $P_{25}$, $P_{50}$, and $P_{75}$ denote the 25th, 50th (median), and 75th percentiles of the severity distribution, respectively.

\begin{figure*}[!htbp]
\centering
\includegraphics[width=0.4\textwidth]{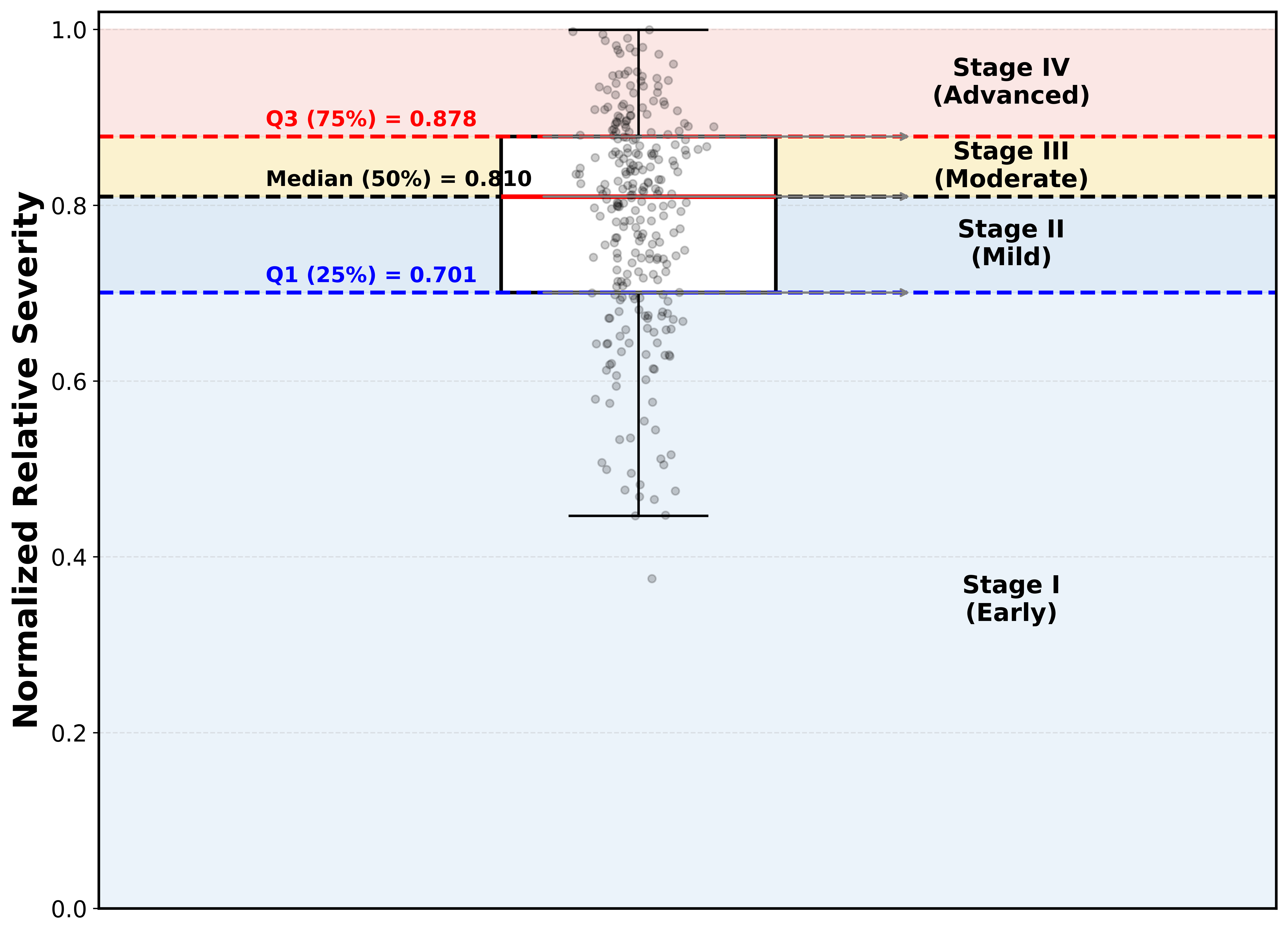}
\vspace{-3mm}
\caption{
Quartile-based disease stage construction employed in AppleScab-LT.}
\label{fig:quartile_distribution}
\end{figure*}

Figure~\ref{fig:quartile_distribution} illustrates the quartile-based disease stage construction strategy, while Table~\ref{tab:stage_statistics} summarizes the corresponding stage-wise statistics. Together, they demonstrate that the proposed partitioning produces four balanced disease stages while preserving the continuous nature of disease progression.

\begin{table}[!htbp]
\centering\small
\caption{Distribution of disease stages generated using quartile-based partitioning of normalized relative disease severity.}
\label{tab:stage_statistics}

\begin{tabular}{lcccc}
\hline
\textbf{Disease Stage} &
\textbf{Severity Range} &
\textbf{Samples} &
\textbf{Relative Severity Range}\\
\hline

Stage I &
$\leq Q_1$ &
66 &
$\leq 0.701$\\

Stage II &
$Q_1-Q_2$ &
66 &
0.7001 – 0.810\\

Stage III &
$Q_2-Q_3$ &
66 &
0.810 – 0.878\\

Stage IV &
$>Q_3$ &
66 &
$> 0.878$\\

\hline
Total &
--
&
264
&
100 &
--\\
\hline
\end{tabular}

\end{table}

The resulting balanced disease stages provide a standardized foundation for temporal disease stage classification while preserving the inherent continuity of disease progression. Consequently, AppleScab-LT supports both continuous severity analysis and discrete stage-based learning, making it suitable for evaluating conventional image classification methods as well as progression-aware temporal learning algorithms.

\subsection{Dataset Diversity Analysis}
\label{sec:Dataset Diversity Analysis}
\begin{figure*}[!htbp]
\centering
\includegraphics[width=0.6\textwidth]{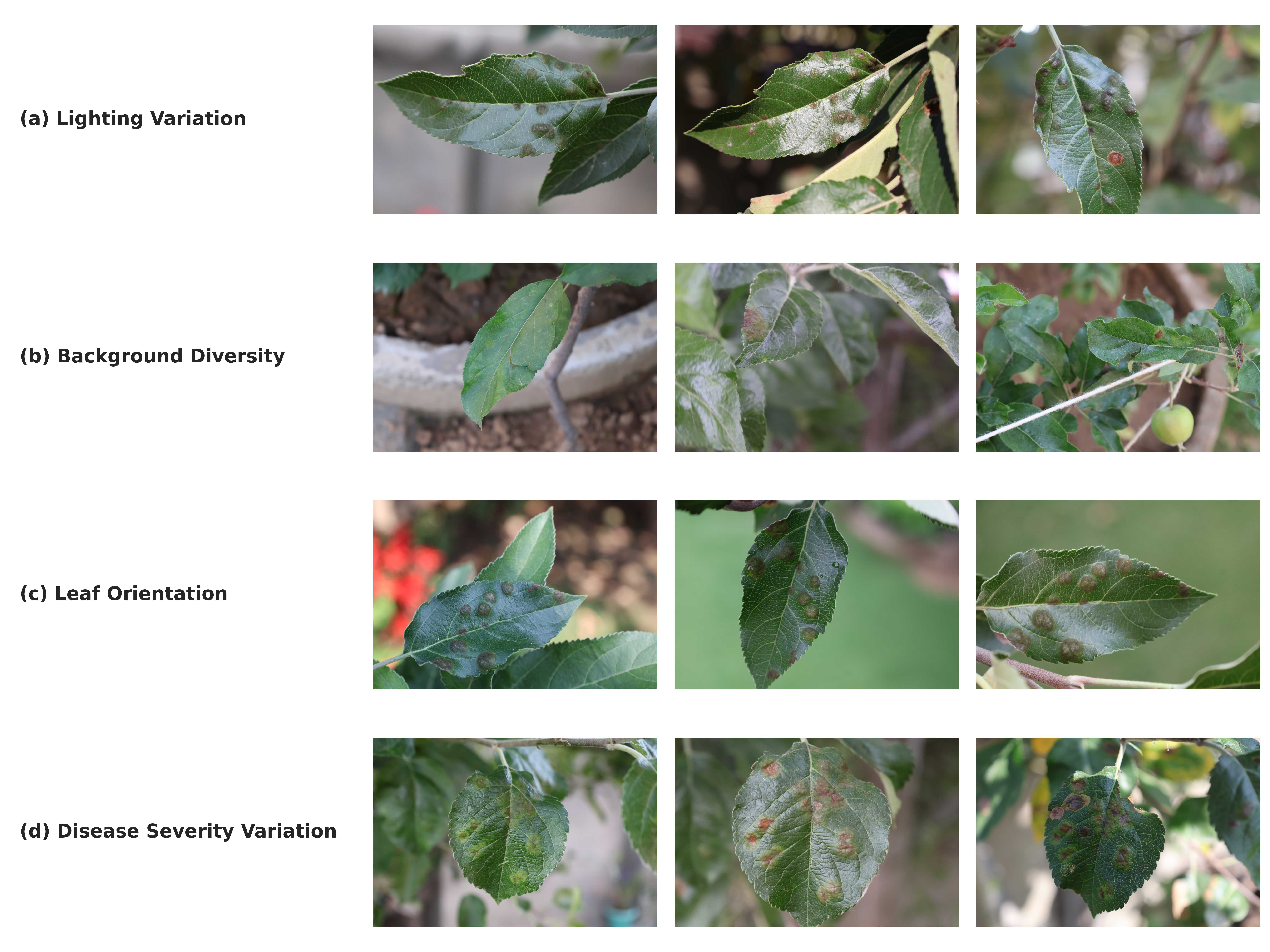}
\vspace{-3mm}
\caption{Representative examples illustrating the visual diversity captured by AppleScab-LT : (a) illumination conditions, (b) background complexity, (c) leaf orientation, and (d) disease severity progression.}
\label{fig:dataset_diversity}
\end{figure*}

A defining characteristic of AppleScab-LT is its ability to capture disease progression under naturally occurring orchard conditions. Consequently, the datset contains substantial variability arising from environmental conditions, biological differences among leaves, and practical challenges encountered during long-term field monitoring. This combination allows to capture both the biological evolution of apple scab and the practical imaging variability encountered during repeated orchard monitoring, thereby providing a more realistic foundation for developing temporal disease intelligence systems. Figure~\ref{fig:dataset_diversity} presents representative examples illustrating the diversity captured within AppleScab-LT. The dataset includes images acquired under varying illumination conditions, background complexity, leaf orientation, and  spectrum of disease development observed during advanced disease progression. 

Unlike laboratory datasets that intentionally minimize environmental variation, AppleScab-LT preserves these naturally occurring acquisition conditions to maintain ecological validity. Consequently, the dataset more faithfully represents real-world orchard scenarios and provides a challenging evaluation platform for disease progression analysis, temporal disease modelling, disease stage classification, and progression-aware artificial intelligence systems.

\begin{tcolorbox}[colback=blue!5!white,
                  colframe=blue!60!black,
                  title=\textbf{Answer to RQ4},
                  fonttitle=\bfseries,
                  fontupper=\footnotesize]

AppleScab-LT captures the essential temporal and visual characteristics required for longitudinal disease intelligence. The dataset preserves repeated observations of the same physical leaves, quantitative disease severity measurements, variable-length temporal sequences, diverse temporal disease descriptors, quartile-based disease stage construction, inter-leaf biological variability, and realistic acquisition variability under natural orchard conditions. Together, these characteristics provide a comprehensive representation of disease evolution, extending beyond conventional static image datasets.
\end{tcolorbox}

\section{RQ5: Dataset Organization and Research Utility}
\label{sec:Dataset Organization and Research Utility}
\subsection{Dataset Organization}
\label{sec:Dataset Organization}
\begin{figure*}[!htbp]
\centering
\includegraphics[width=0.6\textwidth]{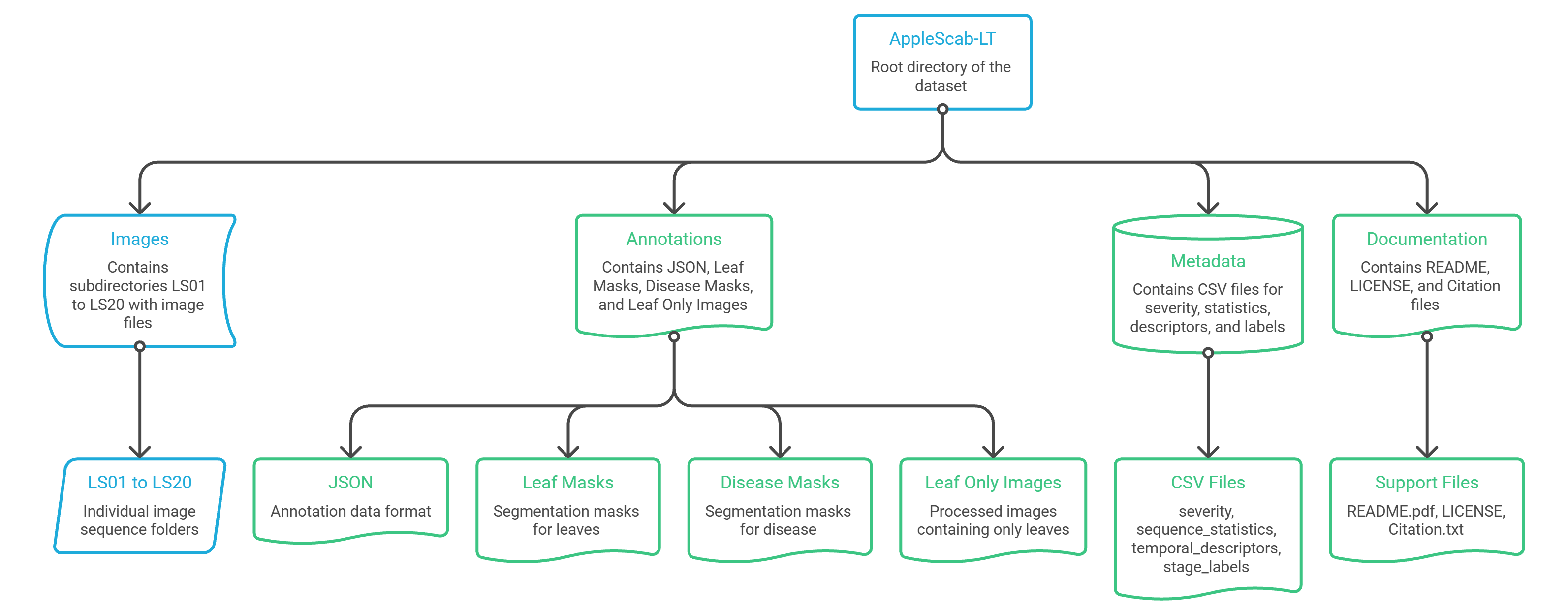}
\vspace{-3mm}
\caption{
Hierarchical organization of the AppleScab-LT dataset.}
\label{fig:dataset_organization}
\end{figure*}

To facilitate reproducibility, interoperability, and long-term usability, AppleScab-LT follows a hierarchical dataset organization designed specifically for longitudinal disease progression analysis. The directory structure separates raw observations, image annotations, segmentation masks, metadata, and temporal measurements while preserving the correspondence among all files belonging to the same leaf sequence. Each monitored leaf sequence is assigned a unique sequence identifier (e.g., LS01--LS20), allowing all observations acquired from the same physical leaf to be grouped together throughout the monitoring period. Within each sequence directory, there is a day-wise directory and within that directory, there are observations containing the original RGB image together with its associated annotation file, segmentation masks, and temporal metadata. This organization enables straightforward retrieval of both image-level and sequence-level information while supporting reproducible temporal analysis. In addition to the image data, AppleScab-LT provides several complementary metadata files describing disease severity, temporal progression, sequence statistics, and acquisition information. Figure~\ref{fig:dataset_organization} illustrates the overall organization of AppleScab-LT.

\subsection{Research Applications}
\label{sec:research_applications}

AppleScab-LT has been designed as a longitudinal dataset that supports a wide range of computer vision, temporal learning, and intelligent agricultural applications. Unlike conventional plant disease datasets that primarily target static image classification, the multimodal organization of AppleScab-LT enables disease analysis across multiple levels, from pixel-level segmentation to long-term temporal forecasting. Table~\ref{tab:research_applications} summarizes the principal research applications supported by the dataset together with their recommended evaluation metrics and intended objectives.

\begin{table*}[!htpb]
\centering
\caption{Research applications supported by AppleScab-LT together with recommended evaluation metrics and corresponding objectives.}
\label{tab:research_applications}
\vspace{-3mm}
\renewcommand{\arraystretch}{0.9}
\small
\resizebox{\textwidth}{!}{
\begin{tabular}{p{3.5cm} p{7.2cm} p{4.8cm} p{6.2cm}}
\hline
\textbf{Research Application} &
\textbf{Dataset Support} &
\textbf{Recommended Metrics} &
\textbf{Research Objective} \\
\hline

Disease Classification &
RGB images with disease stage labels support binary, multiclass, and progression-aware disease recognition. &
Accuracy, Precision, Recall, F1-score, Cohen's $\kappa$, ROC-AUC &
Evaluate disease recognition performance and stage-wise classification accuracy. \\

Semantic Segmentation &
Polygon annotations together with binary leaf and lesion masks enable precise lesion localization and disease area estimation. &
IoU, Dice Coefficient, Precision, Recall &
Assess spatial agreement between predicted and reference lesion masks. \\

Disease Severity Estimation &
Pixel severity, colour intensity severity, and normalized relative severity measurements support quantitative disease assessment. &
MAE, RMSE, $R^2$, Pearson Correlation &
Measure agreement between predicted and reference severity values. \\

Disease Progression Modelling &
Longitudinal severity measurements enable modelling of continuous disease development over time. &
MAE, RMSE, $R^2$, Dynamic Time Warping (DTW) &
Evaluate continuous disease trajectory prediction throughout disease progression. \\

Disease Progression Forecasting &
Repeated temporal observations facilitate forecasting of future disease growth and stage transitions. &
MAE, RMSE, $R^2$, Dynamic Time Warping (DTW) &
Assess temporal prediction accuracy and progression forecasting capability. \\

Longitudinal Disease Tracking &
Unique sequence identifiers maintain correspondence of individual leaves throughout the monitoring period. &
Track Purity, Identity Preservation Rate, Sequence Consistency &
Evaluate temporal consistency and preservation of leaf identity across observations. \\

Digital Twin Development &
Repeated visual observations and quantitative disease descriptors provide structured temporal information for virtual leaf and orchard modelling. &
Simulation Error, MAE, RMSE, Temporal Consistency &
Support virtual disease simulation, monitoring, and precision orchard management. \\

Representation and Self-Supervised Learning &
Diverse visual conditions, temporal observations, segmentation masks, and structured metadata support representation learning and transfer learning. &
Linear Probe Accuracy, Few-shot Accuracy, Transfer Learning Performance &
Assess representation quality for downstream agricultural vision applications. \\

\hline
\end{tabular}}
\end{table*}

\subsection{Recommended Evaluation Protocol}
\label{sec:Recommended Evaluation Protocol}

In longitudinal disease monitoring, multiple observations correspond to the same physical leaf acquired at different stages of disease development. Consequently, randomly assigning individual images to different dataset partitions would allow visually similar observations from the same leaf to appear simultaneously in both the training and testing sets. Such temporal information leakage leads to overly optimistic performance estimates and fails to evaluate the true generalization capability of disease progression models. To avoid this problem, AppleScab-LT recommends a Leave-One-Leaf-Out Cross-
Validation (LOLOCV) protocol that treats each monitored leaf sequence as an independent experimental unit. Under this protocol, all observations belonging to a single leaf sequence are assigned exclusively to one dataset partition. During each validation iteration, one complete leaf sequence is reserved for testing, one sequence is used for validation, and the remaining leaf sequences are used for model training. The procedure is repeated until every leaf sequence has served as the testing set exactly once. The LOLOCV protocol was selected because it maximizes the utilization of the limited longitudinal observations while simultaneously providing an unbiased estimate of model generalization to previously unseen leaves. 

\begin{figure*}[!htbp]
\centering
\includegraphics[width=0.3\textwidth]{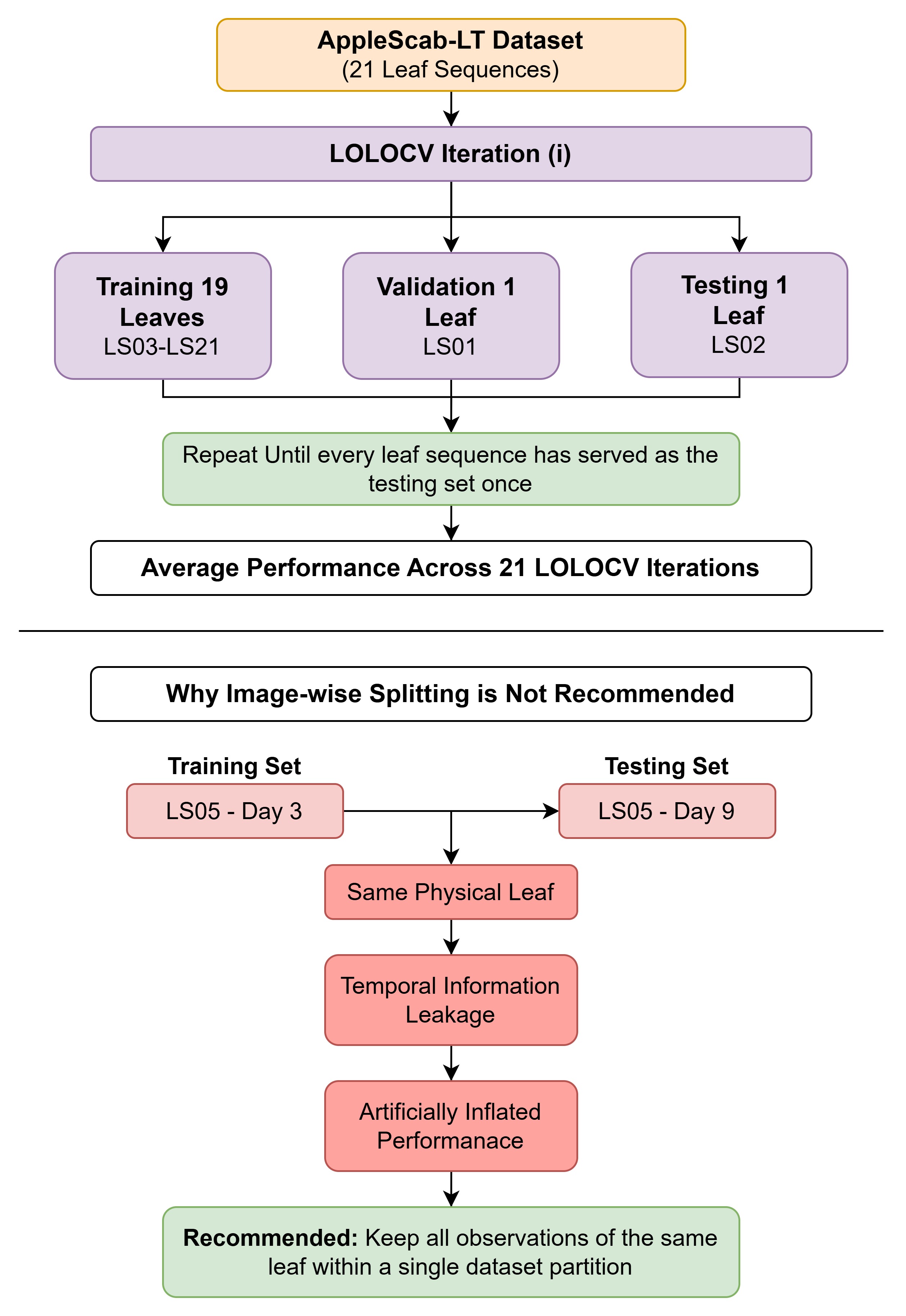}
\vspace{-3mm}
\caption{
Recommended Leave-One-Leaf-Out Cross-Validation (LOLOCV) protocol for AppleScab-LT.
}
\label{fig:LOLOCV protocol}
\end{figure*}

Figure~\ref{fig:LOLOCV protocol} shows the recommended LOLOCV protocol. By preserving complete longitudinal sequences throughout the evaluation process, the protocol provides an unbiased estimate of model performance on previously unseen leaves while fully exploiting the limited number of longitudinal observations available in the dataset. Maintaining a standardized leaf-wise evaluation strategy ensures that reported performance differences reflect genuine methodological improvements rather than inconsistencies in dataset partitioning or experimental design. Unlike conventional image-wise cross-validation protocols, LOLOCV explicitly treats each monitored leaf as the independent biological unit, thereby preserving the temporal coherence of disease progression throughout model evaluation.

\subsection{Reference Implementations}
\label{sec:reference_implementations}
To demonstrate the practical applicability of AppleScab-LT, two representative temporal deep learning architectures were evaluated using the recommended Leave-One-Leaf-Out Cross-Validation (LOLOCV) protocol. These implementations are intended solely as reference examples rather than state-of-the-art models, enabling the validation of experimental pipelines and the comparison of future approaches under a consistent evaluation setting. Table~\ref{tab:baseline_results} summarizes their representative performance. The selected reference implementations comprise a conventional convolutional recurrent architecture (CNN--LSTM) and a transformer-based recurrent architecture (Vision Transformer (ViT)--LSTM). Both models process longitudinal image sequences by extracting spatial representations from individual observations, followed by an LSTM network that models temporal dependencies across successive observations. Table~\ref{tab:baseline_results} summarizes the representative performance ranges obtained using the recommended LOLOCV protocol. These results are intended to provide approximate reference values for future comparison rather than definitive performance limits of AppleScab-LT. Since model performance depends upon implementation details, hyperparameter selection, and training strategies, future methods may reasonably achieve higher accuracies while using the same evaluation protocol. The reference implementations further demonstrate that AppleScab-LT supports modern temporal deep learning approaches capable of exploiting disease progression information preserved within longitudinal leaf sequences.

\begin{table}[!htbp]
\centering
\caption{Representative reference implementation results on AppleScab-LT.}
\label{tab:baseline_results}
\renewcommand{\arraystretch}{0.9}
\small
\begin{tabular}{lcccc}
\hline
\textbf{Model} &
\textbf{Spatial Encoder} &
\textbf{Temporal Module} &
\textbf{Accuracy Range} &
\textbf{Macro F1 Range} \\
\hline

CNN--LSTM &
CNN &
LSTM &
79--84\% &
0.79--0.84 \\

ViT--LSTM &
Vision Transformer &
LSTM &
81--86\% &
0.81--0.86 \\

\hline
\end{tabular}
\end{table}

\begin{tcolorbox}[colback=blue!5!white,
                  colframe=blue!60!black,
                  title=\textbf{Answer to RQ5},
                  fonttitle=\bfseries,
                  fontupper=\footnotesize]

AppleScab-LT provides a standardized longitudinal resource comprising a hierarchical dataset organization, comprehensive metadata, representative research applications, a recommended leaf-wise evaluation strategy, and reference implementations. Together, these components support reproducible research on disease classification, lesion segmentation, severity estimation, disease progression modelling, longitudinal tracking, and temporal disease forecasting.

\end{tcolorbox}

\section{Discussion}
\label{sec:discussion}
The development of AppleScab-LT was motivated by the observation that existing plant disease datasets are predominantly designed for static image classification and therefore provide limited support for studying disease evolution over time. Throughout this work, five research questions were formulated to guide the design, construction, validation, and characterization of this dataset. This discussion synthesizes the principal findings by revisiting each research question and highlighting the broader implications of AppleScab-LT for temporal disease intelligence and precision agriculture.

\subsection{RQ1: Existing Plant Disease Datasets}
\label{sec:}

The systematic review and comparative analysis reveal that existing plant disease datasets have primarily been developed for static disease recognition rather than longitudinal disease analysis. Most publicly available datasets consist of isolated images acquired at a single time point, treating each image as an independent sample without preserving temporal continuity between repeated observations of the same leaf. While these datasets have significantly advanced image-based disease classification and detection, they provide limited support for analysing disease evolution over time. The lack of temporal information restricts applications such as disease progression modelling, stage transition analysis, growth-rate estimation, and disease forecasting. In addition, most available datasets do not include repeated observations, quantitative severity measurements, longitudinal annotations, or standardized temporal evaluation protocols. These findings highlight a clear gap between conventional plant disease datasets and the requirements of temporal disease analysis, motivating the development of AppleScab-LT.

\subsection{RQ2: Development of a Longitudinal Field Dataset}
\label{sec:}

AppleScab-LT demonstrates that reliable longitudinal disease datasets can be developed through systematic field monitoring and standardized acquisition protocols. Instead of collecting isolated disease images, individual apple leaves were repeatedly monitored throughout disease development under natural orchard conditions, preserving the temporal relationships required for progression analysis. The longitudinal dataset emerged naturally from a broader multi-disease apple image collection, reflecting practical field acquisition where infected leaves became available at different stages of the growing season. Combining high-resolution RGB images with polygon annotations, binary masks, quantitative severity measurements, and structured metadata further enhances the dataset beyond conventional image repositories while maintaining ecological validity under real orchard conditions.

\subsection{RQ3: Reliability of Longitudinal Annotation}
\label{sec:}

Reliable longitudinal datasets require stronger quality assurance than conventional image datasets because annotation errors may propagate across entire temporal sequences. AppleScab-LT addresses this challenge through manual polygon annotation, expert verification, automated quality assurance, and temporal consistency validation. Beyond image-level verification, the proposed framework ensures consistent leaf identity across repeated observations while preserving complete sequence integrity throughout the monitoring period. Automated validation further minimizes missing annotations, incorrect labels, file inconsistencies, and sequence-level errors, resulting in a reliable dataset suitable for both image-level and temporal disease analysis.

\subsection{RQ4: Temporal Characterization of Disease Progression}
\label{sec:}

The dataset characterization demonstrates that AppleScab-LT preserves substantially richer information than conventional disease image datasets. In addition to repeated observations, it includes quantitative severity measurements, temporal descriptors, disease stage labels, inter-leaf variability, progression dynamics, and realistic acquisition variability under natural orchard conditions. The analysis further shows that disease progression follows diverse temporal trajectories rather than a single homogeneous pattern, reflecting the biological variability of apple scab development. Furthermore, the quartile-based disease stage construction provides an objective mechanism for converting continuous disease severity into balanced discrete stages, supporting both severity estimation and disease stage classification.

\subsection{RQ5: AppleScab-LT as a Standardized Longitudinal Dataset}
\label{sec:}

Beyond dataset construction, this work provides a standardized resource for longitudinal plant disease analysis through a well-defined directory structure, comprehensive metadata, evaluation protocol, supported tasks, and reference implementations. The proposed Leave-One-Leaf-Out Cross-Validation (LOLOCV) protocol prevents temporal information leakage between training and testing data, providing a more reliable assessment of model generalization to previously unseen leaf sequences. In addition, the availability of RGB images, segmentation masks, quantitative severity measurements, disease stage labels, temporal descriptors, and structured metadata supports a wide range of applications, including disease classification, segmentation, progression modelling, forecasting, digital twins, self-supervised learning, and temporal disease intelligence.

The five research questions formulated at the beginning of this study have been systematically addressed through the development and evaluation of AppleScab-LT. Collectively, these contributions transform AppleScab-LT from a conventional image collection into a comprehensive longitudinal dataset that supports image-level analysis, disease progression modelling, temporal forecasting, and future progression-aware artificial intelligence.

\section{Limitations and Future Work}
\label{sec:limitations}

Although AppleScab-LT provides a comprehensive longitudinal resource for apple scab disease progression analysis, the current release reflects the scope of the present data collection campaign while highlighting opportunities for continued expansion.

AppleScab-LT focuses exclusively on apple scab (\textit{Venturia inaequalis}) collected from orchards in the Kashmir region under natural field conditions. This disease-specific design enables detailed temporal analysis, although incorporating additional sequences and other diseases would further improve generalizability. Longitudinal monitoring also requires repeated acquisition of individually tracked leaves over extended periods, making data collection substantially more demanding than conventional image datasets. In addition, manual polygon annotation of leaf and lesion regions, followed by expert verification and quality assurance, represents a resource-intensive process that limits rapid dataset growth.

AppleScab-LT is designed as an extensible longitudinal resource. Planned expansions include increasing the number of monitored leaf sequences, extending observation periods, incorporating additional diseases such as powdery mildew, alternaria leaf blotch, and naturally occurring multi-disease interactions.

The dataset also provides a foundation for integrating complementary sensing modalities. The extensions can broaden its applicability to disease forecasting, precision orchard management, and temporal disease intelligence. Despite its current scope, AppleScab-LT represents an organized longitudinal datasets for apple scab progression under real-field orchard conditions, providing a scalable foundation for temporal plant disease analysis.

\section{Conclusion}
\label{sec:conclusion}

This paper introduced AppleScab-LT, a longitudinal dataset developed for temporal analysis of apple scab progression under real-field orchard conditions. Unlike conventional plant disease datasets that primarily consist of static images, AppleScab-LT preserves the temporal evolution of disease through repeated observations of the same physical leaves, providing a valuable resource for studying disease dynamics over time. The dataset was developed through systematic field monitoring, expert-guided polygon annotation, rigorous quality assurance, and comprehensive reliability assessment. It comprises high-resolution RGB images, binary segmentation masks, structured annotations, disease severity measurements, temporal descriptors, disease stage labels, and standardized metadata, enabling reproducible research across multiple computer vision and temporal learning tasks. Comprehensive characterization demonstrates that AppleScab-LT captures diverse disease progression patterns, biological variability, balanced disease stages, and realistic field conditions. In addition, the dataset provides a standardized organization, a leaf-wise evaluation protocol, and reference implementations to facilitate consistent and reproducible model evaluation. Collectively, these contributions establish AppleScab-LT as a valuable resource for advancing temporal plant disease modelling, progression-aware artificial intelligence, and decision-support systems for precision agriculture.

\section*{Data Availability}
The dataset presented in this study will be shared by the corresponding author upon reasonable request.

\section*{Funding}
The authors received no financial support for the research, authorship, and/or publication of this article.

\section*{Competing Interests}
The authors declare that they have no competing interests.
{
\footnotesize
\bibliographystyle{unsrtnat}
\bibliography{mybib}
}

\end{document}